\documentclass{article}
\usepackage{iclr2027_conference,times}

\usepackage{url}
\usepackage{booktabs}
\usepackage{arydshln}
\usepackage{amsmath}
\usepackage{amssymb}
\usepackage{multirow}
\usepackage{graphicx}
\usepackage{xcolor}
\usepackage{wrapfig}
\usepackage{float}
\usepackage{enumitem}
\usepackage{abbrv}
\usepackage{etoolbox}
\usepackage{hyperref}
\hypersetup{hidelinks}

\AtEndPreamble{
    \usepackage[capitalize]{cleveref}
    \crefname{section}{Sec.}{Secs.}
    \Crefname{section}{Section}{Sections}
    \crefname{table}{Tab.}{Tabs.}
    \Crefname{table}{Table}{Tables}
}

\newcommand{\customparagraph}[1]{\par\noindent\textbf{#1:}}

\title{DewHand: Direct Metric World-Space Hand Motion Reconstruction from Egocentric Videos}

\author{\textbf{Naru Suzuki}$^{1,3}$ \quad \textbf{Satoshi Ikehata}$^{2}$ \quad \textbf{Takehiko Ohkawa}$^{1}$\\
\textbf{Yoichi Sato}$^{1}$ \quad \textbf{Shuhei Kurita}$^{2,3}$\\[0.3em]
 $^{1}$The University of Tokyo \quad $^{2}$National Institute of Informatics  \quad $^{3}$NII LLMC}

\begin{document}

\maketitle

\begin{figure}[h]
\centering
\includegraphics[width=\linewidth,height=0.27\textheight,keepaspectratio]{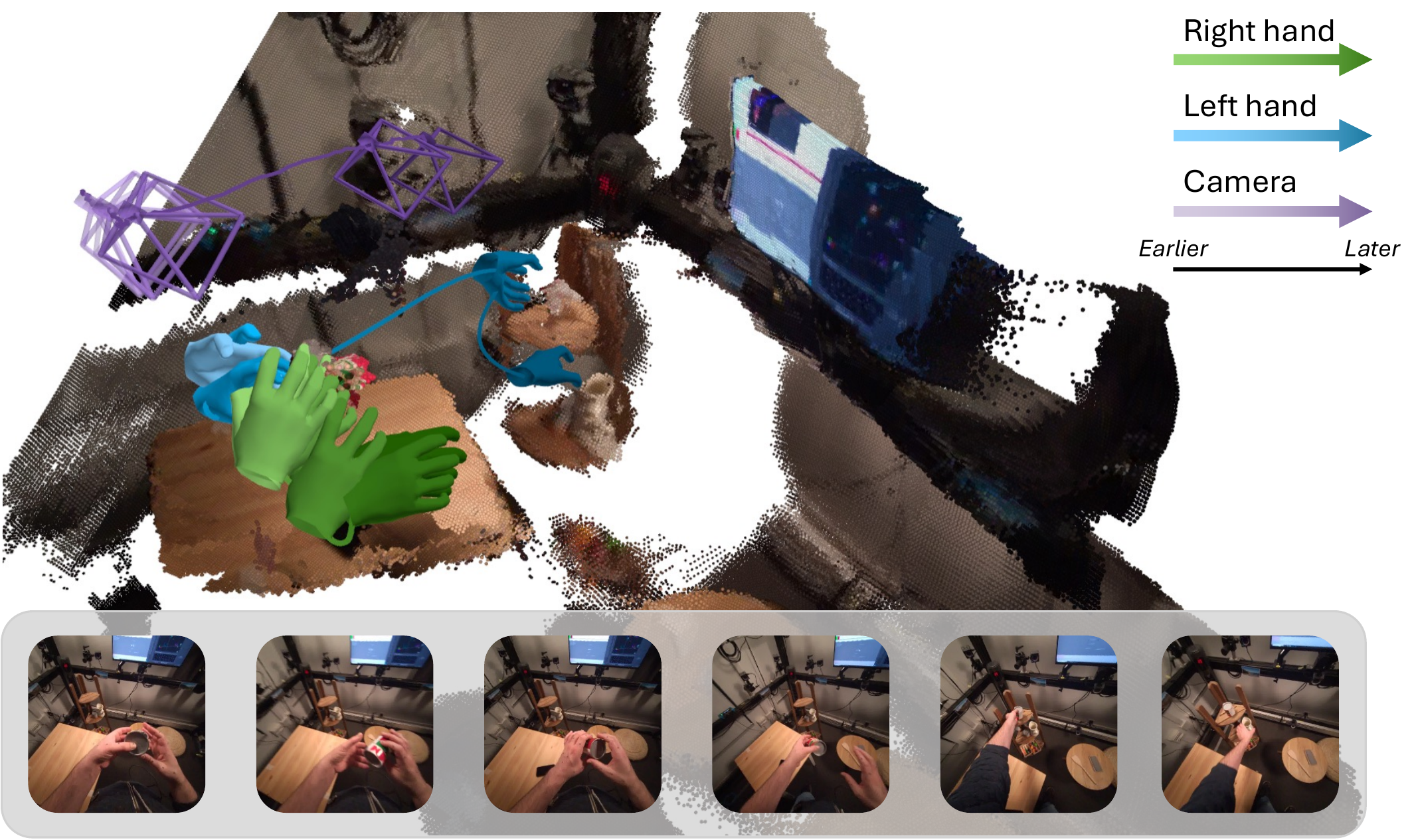}
\vspace{-5pt}
\caption{\textbf{DewHand} directly reconstructs the motion of both hands and the camera in a shared metric world coordinate system from monocular egocentric video. See more on \href{https://narusuzuki.github.io/projects/26-dewhand/}{\textcolor[HTML]{EC008B}{\mbox{the project page}}}.}
\label{fig:teaser}
\end{figure}

\begin{abstract}
Recovering metric world-space hand motion from egocentric video supports understanding human interactions and transferring them to embodied agents. Three challenges complicate this task: multi-stage pipelines propagate errors from independently estimated hand and camera motion; monocular observations leave physical scale ambiguous; and occlusion and out-of-view hands admit multiple plausible motion trajectories. We present \textbf{DewHand}, a unified model for direct joint reconstruction of world-space hand and camera motion.
To avoid compounding errors from multi-stage estimation, DewHand jointly predicts hand and camera motion in a shared world coordinate system within a single-stage pipeline. To resolve metric-scale ambiguity, it combines scene-level geometric features from VGGT-$\Omega$ with metric-space supervision, while hand-specific features preserve fine-grained articulation. To handle occlusion and out-of-view hands, we formulate reconstruction as conditional motion generation with flow matching. Under the same architecture, this generative formulation outperforms direct regression in both accuracy and temporal smoothness with only two sampling steps.
Experiments on HOT3D and ARCTIC demonstrate improved world-space accuracy and temporal consistency over state-of-the-art methods. DewHand reduces rigid-aligned world-space error relative to StableHand from $32.9$ to $28.4$\,mm on HOT3D and from $76.2$ to $36.2$\,mm on ARCTIC. It also recovers camera scale within $10\%$ on two thirds of HOT3D clips, versus about a quarter for SLAM. Moreover, in zero-shot evaluations on HoloAssist, H2O, and TACO, captured with different devices, DewHand achieves the lowest rigid-aligned world-space error (WA-MPJPE) on each dataset.
\end{abstract}

\clearpage
\section{Introduction}
\begin{wrapfigure}{r}{0.55\textwidth}
\vspace{-\intextsep}
\centering
\includegraphics[width=\linewidth]{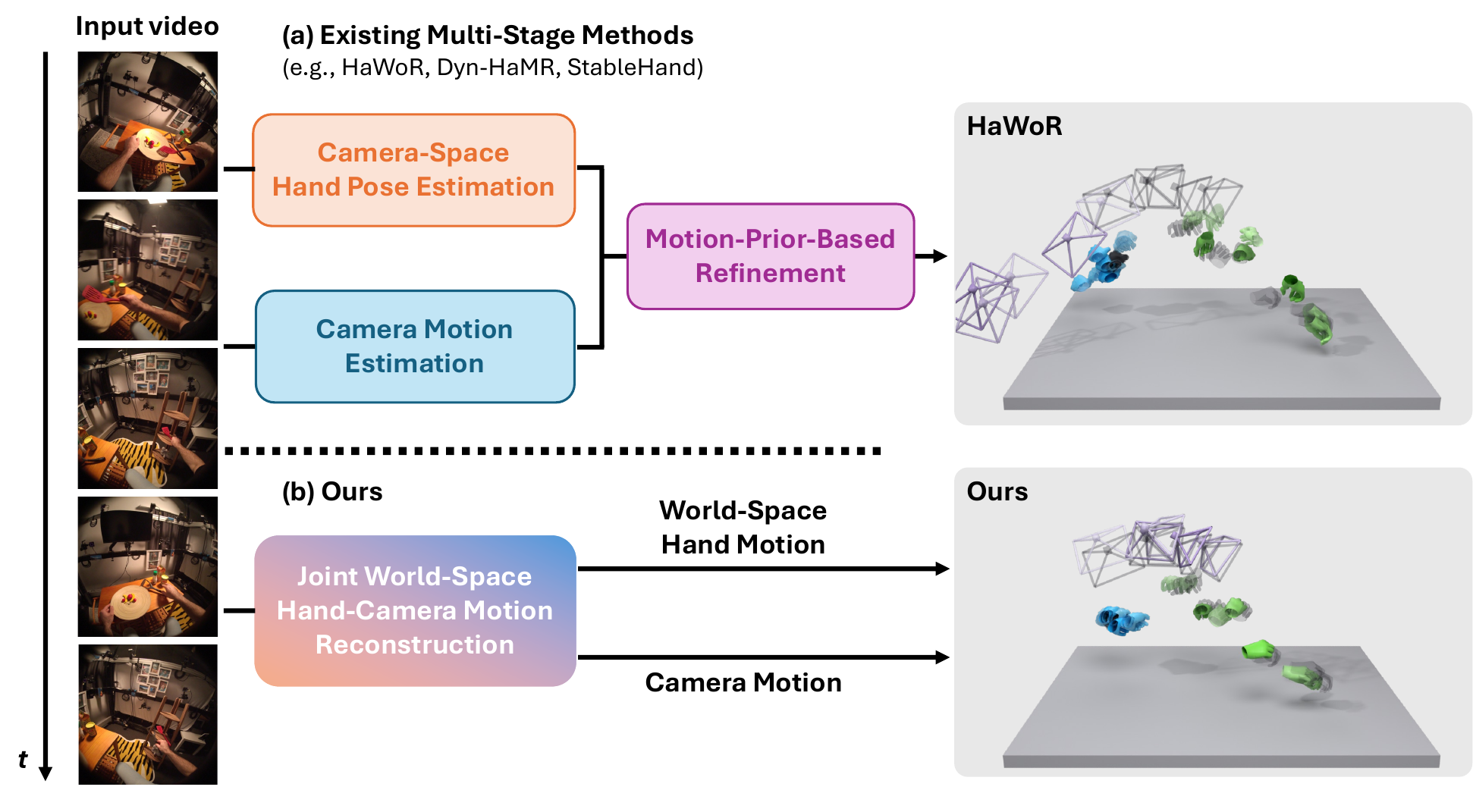}
\vspace{-10pt}
\caption{\textbf{Direct world-space hand-camera generation.}
(a) Prior pipelines~\citep{hawor,dynhamr,stablehand} combine separate hand and camera estimates.
(b) DewHand jointly generates both motions in metric world space.
GT is overlaid in translucent dark gray on each reconstruction; predictions use green for the right hand,
blue for the left hand, and purple for the camera.}
\label{fig:pipeline_comparison}
\end{wrapfigure}

Hands are our primary means of interacting with the physical world. Recovering their 3D motion
from wearable cameras can support the analysis of human activity, the capture of manipulation
skills, and the transfer of human demonstrations to embodied agents. These applications require
more than an accurate hand pose in each video frame: they require a metrically scaled trajectory in
a stable world coordinate system, so that a hand remains correctly placed as both the wearer and the
camera move.

This work addresses world-space 3D hand motion reconstruction from monocular egocentric video.
Given a video, the goal is to recover the time-varying pose of both hands in a common world frame.
Consider a wearer reaching for an object while turning their head. In camera coordinates, even a
stationary hand appears to move with the camera; conversely, camera motion can hide or amplify the
hand's motion in the image. A useful reconstruction must separate these effects and preserve the
hand's position relative to the surrounding scene throughout the interaction.

Three challenges motivate our formulation. First, accurate local hand reconstruction does not by
itself ensure a consistent world-space trajectory. Despite strong per-frame estimators~\citep{hamer,wilor},
existing world-space pipelines typically estimate hands and camera motion separately, compose their
outputs, and optionally apply temporal optimization or motion-prior refinement~\citep{hawor,dynhamr,handflow,stablehand}.
Camera drift and mismatches between these estimates can therefore propagate into the reconstructed
hand motion. Second, monocular video provides projective observations rather than direct metric
measurements, leaving physical scale ambiguous even when local articulation is accurately recovered.
Third, self-occlusion, hand-object interactions, motion blur, and frequent exits from the field of
view make hand observations incomplete. The visible evidence may be consistent with multiple motion
trajectories, motivating a conditional distribution over hand and camera motion.

We introduce \textbf{DewHand}, a unified model that directly generates metric world-space
motion of both hands and the camera from egocentric video. To avoid error propagation, we jointly
generate wrist poses, local articulation, and camera motion in a shared world frame instead of
composing independent estimates. To support
metric-scale recovery, we condition on geometric features from the frozen VGGT-$\Omega$
backbone~\citep{vggtomega} and supervise in metric space; hand-specific features add cues for local articulation. To handle ambiguous and missing observations, we learn a conditional
distribution over joint hand-camera motion using flow matching. Compared with regressing a single trajectory, this generative
formulation produces more accurate and temporally smoother motion, while exhibiting greater sample
diversity where the hand is unobserved.

Our contributions are threefold:
\begin{itemize}
\item We formulate egocentric world-space hand reconstruction as direct joint generation of metric
dual-hand and camera motion instead of composing separately estimated hands and camera;
supervising the generated camera reduces rigid-aligned hand error by $6.4$\,mm.

\item We develop a flow-matching architecture that combines geometry-aware and hand-specific video
features to jointly model wrist placement, articulation, and camera motion in a shared world frame;
replacing the geometric features with DINOv2 increases hand error by $13$\,mm and camera error by $57$\,mm.

\item We establish state-of-the-art results on HOT3D and ARCTIC, reducing rigid-aligned world error
against StableHand from $32.9$ to $28.4$\,mm and from $76.2$ to $36.2$\,mm, and recovering camera scale
within $10\%$ on two thirds of HOT3D clips, versus about a quarter for SLAM.
\end{itemize}

\section{Related Work}
\customparagraph{3D Hand Pose and Mesh Reconstruction}
Large-scale models such as HaMeR~\citep{hamer} recover detailed articulated pose in the wild,
while WiLoR~\citep{wilor} additionally targets robust 3D hand localization. Interaction
and physical priors~\citep{interprior,affordance,intuitivephysics} and spatio-temporal
context~\citep{sthand} further address occlusion and motion ambiguity. Recent work has explored
complementary strategies for monocular hand reconstruction, including contrastive pre-training in
SiMHand~\citep{Lin2025SiMHand}, controllable synthetic data generation in
SesaHand~\citep{zhao2026sesahand}, exemplar-based in-context reconstruction in
EgoHandICL~\citep{egohandicl}, and kinematics-aware architectures in Rodrigues
Network~\citep{zhang2026rodrinet}. Egocentric datasets and benchmarks capture hand-object
interactions~\citep{ohkawa2023AssemblyHands,fan2024benchmarks}. HOPformer~\citep{hopformer}
models egocentric hand-object pose, and EgoForce~\citep{egoforce} estimates camera-space hand pose.
These advances improve hand pose and shape, but a shared metric world trajectory also requires
camera motion.

\customparagraph{World-Space Hand Motion Reconstruction}
For full-body motion, WHAC~\citep{whac} jointly recovers metric human and camera trajectories.
For articulated hands, HaWoR~\citep{hawor} and
Dyn-HaMR~\citep{dynhamr} construct global motion from camera-space hand estimates and separately
recovered camera trajectories. HandFlow~\citep{handflow} learns camera-space hand motion distributions and
transforms its predictions into world coordinates using SLAM. StableHand~\citep{stablehand} instead
learns world-space refinement of hand observations constructed from separate hand and camera estimates,
using quality-aware initialization to preserve reliable observations. UniHand~\citep{unihand}
generates hand motion through latent diffusion in a fixed coordinate system defined by the first
camera frame, but does not generate the camera trajectory. ReViV~\citep{reviv} jointly models camera
trajectories and camera-space hand kinematics, then aligns the reconstructed viewer and scene in a
metric frame. DewHand jointly generates metric hand and camera trajectories in a single stage, conditioned on geometric and hand-specific features.

\section{Method}
\begin{figure}[t]
\centering
\includegraphics[width=\linewidth]{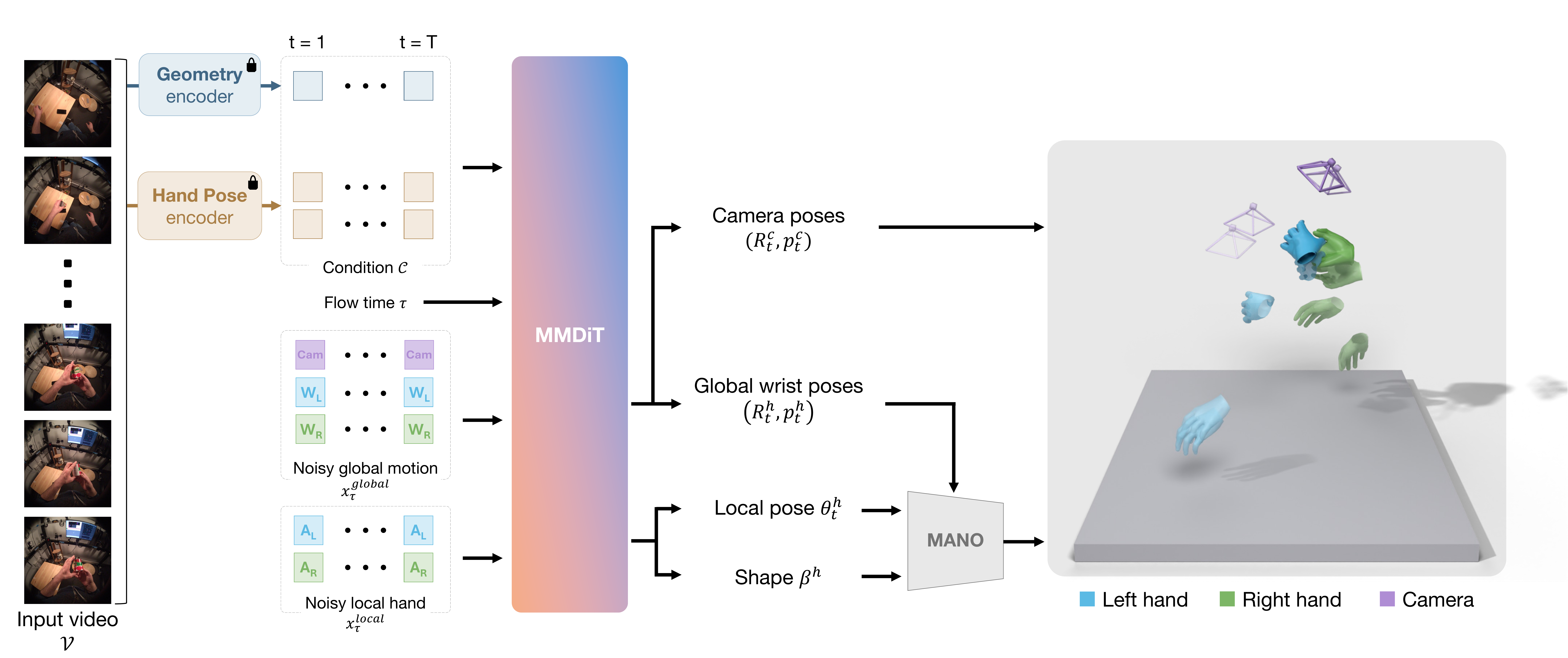}
\vspace{-20pt}
\caption{\textbf{DewHand architecture.} Frozen encoders provide
$\mathcal C=(\mathcal C_{\mathrm{geo}},\mathcal C_{\mathrm{hand}})$.
At flow time $\tau$, the model first applies spatio-temporal attention within each stream, then
joint attention across the global motion, local hand, and condition streams at each frame.
The Multimodal Diffusion Transformer (MMDiT) denoises
$\mathbf{x}^{\mathrm{global}}_{\tau}$ and $\mathbf{x}^{\mathrm{local}}_{\tau}$.
The outputs are camera pose $(\mathbf{R}^c_t,\mathbf{p}^c_t)$ and, for each hand
$h\in\{L,R\}$, global wrist pose $(\mathbf{R}^h_t,\mathbf{p}^h_t)$, local pose
$\boldsymbol{\theta}^h_t$, and shape $\boldsymbol{\beta}^h$.
$\mathrm{Cam}$, $W_h$, and $A_h$ are the corresponding tokens; MANO converts the hand outputs to meshes.}
\label{fig:arch}
\end{figure}

For a monocular egocentric video $\mathcal{V}=\{I_t\}_{t=1}^{T}$, $t$ indexes video frames
and $\tau\in[0,1]$ denotes flow time. Fig.~\ref{fig:arch} summarizes our design. Unlike
HandFlow~\citep{handflow}, which models
camera-space hand motion before SLAM-based lifting, and StableHand~\citep{stablehand}, which
refines world-space observations assembled from separate hand and camera estimates, DewHand
generates both trajectories in a metric world frame without an initial reconstruction.
Geometric conditioning resolves scale ambiguity, and conditional flow modeling handles
missing observations.

\subsection{Problem Formulation and State Representation}
Our goal is to recover the metric 3D motion of both hands \emph{and} the camera in a common world frame.
We take the camera coordinate system of the first frame as the world frame, so that no external
localization, gravity estimate, or scene template is required at inference time.
The outputs in Fig.~\ref{fig:arch} are the camera pose
$(\mathbf{R}^c_t,\mathbf{p}^c_t)\in SO(3)\times\mathbb{R}^3$ and, for each hand
$h\in\{L,R\}$, global wrist pose $(\mathbf{R}^h_t,\mathbf{p}^h_t)\in SO(3)\times\mathbb{R}^3$,
local pose $\boldsymbol{\theta}^h_t\in\mathbb{R}^{15\times 6}$, and clip-level shape
$\boldsymbol{\beta}^h\in\mathbb{R}^{10}$~\mbox{\citep{mano}}. The local pose stores 6D rotations of
15 joints, converted to MANO's 45D axis-angle pose for mesh decoding.
We group the quantities that live in the world frame into a
\emph{global motion stream} $\mathbf{x}^{\mathrm{global}}_t=(\mathbf{R}^L_t,\mathbf{p}^L_t,\mathbf{R}^R_t,\mathbf{p}^R_t,\mathbf{R}^c_t,\mathbf{p}^c_t)$
and the hand-local parameters into a \emph{local hand stream}
$\mathbf{x}^{\mathrm{local}}_t=(\boldsymbol{\theta}^L_t,\boldsymbol{\beta}^L,\boldsymbol{\theta}^R_t,\boldsymbol{\beta}^R)$.
The clip-level shape target is repeated across frames in this stream.
This separation reflects the different geometry and conditioning cues of global rigid motion and
local articulation, while the two streams remain coupled throughout the model. Rotations are
parameterized by the 6D continuous representation~\citep{zhou2019continuity}.

Hands are frequently outside the field of view, occluded, or missed by per-frame detectors, so the
observations can admit multiple plausible trajectories. A deterministic regressor can collapse these
possibilities into an averaged motion. Let $\mathcal{C}=(\mathcal{C}_{\mathrm{geo}},\mathcal{C}_{\mathrm{hand}})$
denote the geometric and hand-specific features extracted from $\mathcal{V}$, as described in
Section~\ref{sec:cond}. Rather than directly regressing a single trajectory, we therefore learn
the conditional motion distribution
$p_{\theta}(\mathbf{x}^{\mathrm{global}}_{1:T},\mathbf{x}^{\mathrm{local}}_{1:T}\mid\mathcal{C})$.
Moreover, apparent hand
motion in egocentric video is induced jointly by hand movement and camera ego-motion. We therefore
generate hands and camera together, allowing the model to separate their contributions while
maintaining their consistency in the shared world frame.
\subsection{Visual Conditioning and Missing Observations}

\label{sec:cond}
Metric world-space reconstruction requires scene-level geometric cues to constrain scale and global
placement, alongside fine-grained hand cues for local articulation. We therefore encode
the video with two complementary frozen backbones: VGGT-$\Omega$~\citep{vggtomega} provides
clip-level scene geometry and camera cues, while HaMeR~\citep{hamer} provides detailed per-hand
appearance features. Together with metric-space supervision (Section~\ref{sec:loss}), the geometric
features support physical-scale recovery without fixing the output to an externally estimated camera
trajectory. VGGT-$\Omega$ jointly processes all $T$ frames, so each per-frame output already
contains temporal context. A learned query attention-pools its patch features. We add a learned
projection of its camera token to the pooled feature and project the sum into one geometry token
$\mathbf{c}^{G}_t$ per frame. HaMeR patch features are spatially pooled and projected into
separate left- and right-hand tokens $\mathbf{c}^{h}_t$, $h\in\{L,R\}$; separate tokens keep a
missing or corrupted observation of one hand from affecting the other.

Hand-specific features are unavailable when a hand is out of view or its detection fails.
Let $m_t^h\in\{0,1\}$ indicate whether a feature is used for hand $h$ at frame $t$ ($1$ if
observed, $0$ if missing or masked during training). We replace unused features with a learned, side-specific
mask token $\mathbf{e}_{\mathrm{mask}}^h$:
\begin{equation}
\widetilde{\mathbf{c}}^h_t
=m_t^h\mathbf{c}^h_t+(1-m_t^h)\mathbf{e}_{\mathrm{mask}}^h.
\end{equation}
The resulting $\widetilde{\mathbf{c}}^h_t$ is the hand-conditioning token passed to the flow model.
During training, we additionally mask individual observations, contiguous spans, and complete hand
tracks. The model must therefore recover missing evidence from the remaining video,
scene, and temporal context, improving robustness to intermittent detector failures. The
condition stream has three tokens per frame: geometry, left hand, and right hand.

\subsection{Joint Hand-Camera Flow Model}
\label{sec:model}
Flow matching~\citep{flowmatching,motionflowmatching} models the trajectories compatible with
incomplete observations by learning transport from noise to data. We apply it to the joint
hand-camera state, conditioned on visual features. The model uses two integration steps at
inference.

At flow time $\tau$, $\mathbf{x}^{\mathrm{global}}_{\tau}$ and
$\mathbf{x}^{\mathrm{local}}_{\tau}$ denote the clip-wide states of the two generated streams.
Here the subscript $\tau$ indexes flow time, while $t$ continues to index video frames within
each state; the subscript $1$ below denotes the clean endpoint of the flow.
We use one token per entity and frame. The global motion stream contains three 9D tokens for the left wrist,
right wrist, and camera. Each token represents 3D translation and 6D rotation in world coordinates. The local hand stream contains two
100D tokens, one per hand, containing the 6D rotations of the 15 articulated MANO joints and the 10D
MANO shape. The condition
stream contains the three tokens defined above. Each input is projected to width $D$, and entity- and
frame-specific embeddings retain its identity and temporal position. The
three streams contain eight tokens in total at every frame.
In Fig.~\ref{fig:arch}, $\mathrm{Cam}$ denotes the camera token, $W_h$ the global wrist
token, and $A_h$ the local hand token for side $h\in\{L,R\}$.

Each layer performs attention in two successive stages: within-stream attention, then joint attention across streams.
\emph{Within-stream attention} first relates the entities at each frame and then aggregates temporal
context per entity. They run independently on the global motion, local hand, and condition
streams, with separate parameters, to build per-stream spatio-temporal representations.

\emph{Joint attention across streams} then lets global motion, local hand, and condition
tokens attend to one another at each frame in a Multimodal Diffusion Transformer (MMDiT)
block~\citep{esser2024scaling}. As in HandFlow~\citep{handflow}, condition tokens
are updated rather than kept fixed.

The flow time $\tau$ and the clip-averaged VGGT camera token provide adaptive modulation throughout the
network. Separate output heads map the final global motion and local hand tokens to clean-trajectory estimates
$\hat{\mathbf{x}}^{\mathrm{global}}_1$ and $\hat{\mathbf{x}}^{\mathrm{local}}_1$. The predicted wrist poses and articulation
are composed through MANO with the clip-averaged shape
$\bar{\boldsymbol{\beta}}^h=T^{-1}\sum_t\hat{\boldsymbol{\beta}}^h_t$, where
$\hat{\boldsymbol{\beta}}^h_t$ is the predicted shape component for hand $h$ at video frame $t$.
The predicted camera entity directly defines the camera trajectory; no
separately estimated camera trajectory is composed with the hands after generation.

\begin{table}[t]\centering
\caption{\textbf{World-space hand motion on HOT3D and ARCTIC.}
Errors in mm, AccEr in $\mathrm{m/s^2}$; wWA is wrist-level WA-MPJPE. Pred.\ shape: estimated MANO shape;
GT shape: privileged ground-truth shape. $\ast$: zero-shot on ARCTIC. $\dagger$: training data
include HaWoR-split evaluation sequences. \textbf{Best}/\underline{second best}:
per split, among rows with the same shape input.
}
\label{tab:main}
\small
\setlength{\tabcolsep}{2pt}

\renewcommand{\arraystretch}{1.12}
\begin{tabular*}{\linewidth}{@{\extracolsep{\fill}}l@{\hspace{5pt}}lrrrrrrrrr@{}}
\toprule
 & & & \multicolumn{2}{c}{W-MPJPE $\downarrow$} & \multicolumn{2}{c}{WA-MPJPE $\downarrow$} & \multicolumn{2}{c}{wWA $\downarrow$} & & \\
\cmidrule(lr){4-5}\cmidrule(lr){6-7}\cmidrule(lr){8-9}
 & Method & PA $\downarrow$ & $\mathrm{Sim}(3)$ & $\mathrm{SE}(3)$ & $\mathrm{Sim}(3)$ & $\mathrm{SE}(3)$ & $\mathrm{Sim}(3)$ & $\mathrm{SE}(3)$ & MRRPE-A $\downarrow$ & AccEr $\downarrow$ \\
\midrule
\multirow{12}{*}{\rotatebox[origin=c]{90}{\textbf{HOT3D}}} & \multicolumn{10}{@{}l@{}}{\textit{HaWoR split}} \\
  & WiLoR\,$+$\,ViPE$^{\dagger}$ & 8.21 & 171.78 & 176.26 & 64.52 & 92.06 & 53.00 & 86.33 & 166.29 & 16.98 \\
 & HaWoR & 7.24 & \underline{151.62} & \underline{162.04} & \underline{55.43} & 84.80 & \underline{46.10} & 79.20 & 138.53 & \underline{14.77} \\
 & HandFlow$^{\dagger}$ & 7.49 & 162.23 & 170.94 & 64.61 & 95.24 & 53.46 & 90.00 & 134.36 & 17.71 \\
 & EgoForce\,$+$\,ViPE$^{\dagger}$ & \underline{7.11} & 158.31 & 167.30 & 60.13 & \underline{79.31} & 49.24 & \underline{73.08} & \underline{114.14} & 15.69 \\
 & HOPformer\,$+$\,ViPE & 13.54 & 223.86 & 231.17 & 85.55 & 106.01 & 60.56 & 85.19 & 199.68 & 20.87 \\
 & \makebox[5.1em][l]{\textbf{Ours}}(Pred. shape) & \textbf{6.76} & \textbf{78.60} & \textbf{77.59} & \textbf{32.27} & \textbf{35.94} & \textbf{24.71} & \textbf{29.49} & \textbf{56.91} & \textbf{1.89} \\
\cmidrule(lr){2-11}
 & \multicolumn{10}{@{}l@{}}{\textit{StableHand split}} \\
 & \makebox[5.1em][l]{StableHand}(Pred. shape) & 11.56 & 107.43 & 117.69 & 37.19 & 43.74 & 23.32 & 32.86 & 55.78 & 3.14 \\
 & \makebox[5.1em][l]{\textbf{Ours}}(Pred. shape) & \textbf{6.25} & \textbf{64.21} & \textbf{63.46} & \textbf{26.45} & \textbf{28.56} & \textbf{20.47} & \textbf{24.07} & \textbf{52.00} & \textbf{2.65} \\
\noalign{\vskip 1pt}\cdashline{2-11}[1pt/1.5pt]\noalign{\vskip 2pt}
 & \makebox[5.1em][l]{StableHand}(GT shape) & \textbf{4.94} & 81.02 & 87.31 & 28.71 & 32.90 & \textbf{20.13} & 26.92 & \textbf{45.35} & 2.92 \\
 & \makebox[5.1em][l]{\textbf{Ours}}(GT shape) & 5.71 & \textbf{63.58} & \textbf{63.33} & \textbf{26.26} & \textbf{28.40} & 20.47 & \textbf{24.08} & 51.97 & \textbf{2.66} \\
\midrule
\multirow{9}{*}{\rotatebox[origin=c]{90}{\textbf{ARCTIC}}} & WiLoR\,$+$\,ViPE & \underline{8.12} & 231.97 & 233.98 & 72.38 & 77.64 & 59.36 & 66.76 & 112.45 & 20.70 \\
 & HaWoR & 12.40 & 311.77 & 339.56 & 94.84 & 140.57 & 74.34 & 124.68 & 191.89 & 24.51 \\
 & HandFlow$^{\ast}$ & 11.01 & 272.98 & 275.82 & 77.62 & 94.70 & 62.08 & 81.96 & 144.62 & 15.14 \\
 & EgoForce\,$+$\,ViPE & 8.88 & 242.60 & 243.37 & 76.83 & 89.12 & 62.39 & 77.33 & 126.72 & 15.79 \\
 & HOPformer\,$+$\,ViPE & 8.36 & \underline{176.74} & \underline{178.65} & \underline{56.13} & \underline{60.65} & \underline{44.34} & \underline{49.43} & \underline{73.94} & 15.96 \\
 & \makebox[5.1em][l]{StableHand}(Pred. shape) & 11.37 & 282.79 & 285.59 & 74.68 & 78.63 & 54.05 & 59.40 & 101.48 & \underline{3.47} \\
 & \makebox[5.1em][l]{\textbf{Ours}}(Pred. shape) & \textbf{8.10} & \textbf{100.05} & \textbf{97.11} & \textbf{33.17} & \textbf{36.35} & \textbf{26.05} & \textbf{29.68} & \textbf{49.34} & \textbf{2.22} \\
\noalign{\vskip 1pt}\cdashline{2-11}[1pt/1.5pt]\noalign{\vskip 2pt}
 & \makebox[5.1em][l]{StableHand}(GT shape) & 10.44 & 282.45 & 282.69 & 72.89 & 76.22 & 52.52 & 56.99 & 97.90 & 3.55 \\
 & \makebox[5.1em][l]{\textbf{Ours}}(GT shape) & \textbf{8.14} & \textbf{99.20} & \textbf{97.00} & \textbf{33.03} & \textbf{36.18} & \textbf{26.02} & \textbf{29.65} & \textbf{49.33} & \textbf{2.22} \\
\bottomrule
\end{tabular*}
\end{table}

\subsection{Flow Matching and Training Objectives}
\label{sec:loss}
For training, $\mathbf{x}_{\tau}=(\mathbf{x}^{\mathrm{global}}_{\tau},
\mathbf{x}^{\mathrm{local}}_{\tau})$ denotes the full clip state at flow time $\tau$;
subscripts $0$ and $1$ denote flow endpoints, not video frames.
Let $\mathbf{x}_1=(\mathbf{x}^{\mathrm{global}}_{1:T},\mathbf{x}^{\mathrm{local}}_{1:T})$ denote the ground-truth trajectory after
per-dimension standardization, and $\mathbf{x}_0\!\sim\!\mathcal{N}(\mathbf{0},\mathbf{I})$ a
noise sample of the same shape. We draw a single flow time $\tau$ per clip from a logit-normal
distribution, share it across both streams so that they are always denoised at a common noise level,
and form $\mathbf{x}_{\tau}=(1-\tau)\,\mathbf{x}_0+\tau\,\mathbf{x}_1$. The network predicts the clean sample
$\hat{\mathbf{x}}_1$ rather than the velocity, which gives the flow-matching term
\begin{equation}
\mathcal{L}_{\mathrm{fm}}=\big\lVert \hat{\mathbf{x}}^{\mathrm{global}}_1-\mathbf{x}^{\mathrm{global}}_1 \big\rVert^2
+\big\lVert \hat{\mathbf{x}}^{\mathrm{local}}_1-\mathbf{x}^{\mathrm{local}}_1 \big\rVert^2 .
\end{equation}
This term is computed in the standardized parameter space. Let $\mathcal{J}(\mathbf{x})$ denote
the world-space 3D hand joints obtained by decoding state $\mathbf{x}$ through MANO, and let
$\mathcal{J}^{\mathrm{gt}}$ denote the ground-truth joints. As in HandFlow~\citep{handflow}, we
supervise the decoded joints, here in metric world space,
$\mathcal{L}_{\mathrm{joint}}=\mathrm{SmoothL1}\!\left(\mathcal{J}(\hat{\mathbf{x}}_1),\,\mathcal{J}^{\mathrm{gt}}\right)$,
over frames with valid hand ground truth. This converts errors in wrist pose and local articulation
into their physical effect on the complete hand. We supervise the camera directly in metric space with
$\mathcal{L}_{\mathrm{cam}}$, a squared error on its position and 6D rotation.
To supervise temporal changes in wrist translation and rotation, let $\mathbf{q}$ denote
world-space reference points fixed to each wrist. We match their first and second differences
along video time $t$ to the ground truth using
$\mathcal{L}_{\mathrm{vel}}=\mathrm{SmoothL1}(\Delta\hat{\mathbf{q}},\Delta\mathbf{q}^{\mathrm{gt}})$ and
$\mathcal{L}_{\mathrm{acc}}=\mathrm{SmoothL1}(\Delta^2\hat{\mathbf{q}},\Delta^2\mathbf{q}^{\mathrm{gt}})$, respectively.
With loss weights $\lambda_j=\lambda_c=1$ and $\lambda_v=\lambda_a=800$, the total objective is
\begin{equation}
\mathcal{L}=\mathcal{L}_{\mathrm{fm}}+\lambda_{j}\mathcal{L}_{\mathrm{joint}}
+\lambda_{c}\mathcal{L}_{\mathrm{cam}}
+\lambda_{v}\mathcal{L}_{\mathrm{vel}}+\lambda_{a}\mathcal{L}_{\mathrm{acc}}.
\end{equation}
The predicted clean endpoint defines a flow field for each stream
$S\in\{\mathrm{global},\mathrm{local}\}$ as
$\mathbf{v}^{S}_{\theta}(\mathbf{x}_{\tau},\tau,\mathcal{C})=
(\hat{\mathbf{x}}^{S}_1(\mathbf{x}_{\tau},\tau,\mathcal{C})-\mathbf{x}^{S}_{\tau})/(1-\tau)$.
We integrate this field with two Euler steps; sample aggregation is detailed in
Appendix~\ref{app:sampling_settings}.

\section{Experiments}
\subsection{World-Space Hand Motion}

\customparagraph{Datasets} 
We evaluate on HOT3D~\citep{hot3d} and ARCTIC~\citep{arctic}, following prior work on world-space
hand motion reconstruction, including HaWoR~\citep{hawor} and HandFlow~\citep{handflow} on HOT3D
and StableHand~\citep{stablehand} on both datasets.
We train on HOT3D (for the StableHand split, a separate model that excludes its test windows;
Appendix~\ref{app:evalsets}) and fine-tune the resulting checkpoint on ARCTIC. For both training and inference, we use sequences of $T=150$ frames at 30\,fps, corresponding to 5-second clips.
\customparagraph{Evaluation Metrics}
We report PA-MPJPE, W-MPJPE, WA-MPJPE, MRRPE-A, and acceleration error (AccEr) on the 21-joint MANO
skeleton. MRRPE-A measures inter-wrist displacement after a shared rigid alignment. We report
WA-MPJPE with similarity ($\mathrm{Sim}(3)$) and rigid ($\mathrm{SE}(3)$) alignment. Rigid alignment
removes differences in world-frame origin and orientation while preserving scale, so scale errors
remain visible; similarity alignment additionally corrects scale. We also report wrist-level WA-MPJPE
(wWA), which scores only the wrist trajectory and thus separates world placement from finger
articulation. Appendix~\ref{app:protocol} details
the scoring protocol. Unless stated otherwise, inference uses two Euler steps and five samples.
\customparagraph{Baselines}
We compare per-frame hand estimators (WiLoR, EgoForce~\citep{egoforce}, HOPformer~\citep{hopformer})
lifted with ViPE, and the video-based methods HaWoR~\citep{hawor}, HandFlow~\citep{handflow}, and
StableHand~\citep{stablehand}; all but HandFlow are trained on ARCTIC. Baseline implementation details are in Appendix~\ref{app:protocol}.

\begin{figure}[t]
\centering
\includegraphics[width=\linewidth,height=0.42\textheight,keepaspectratio]{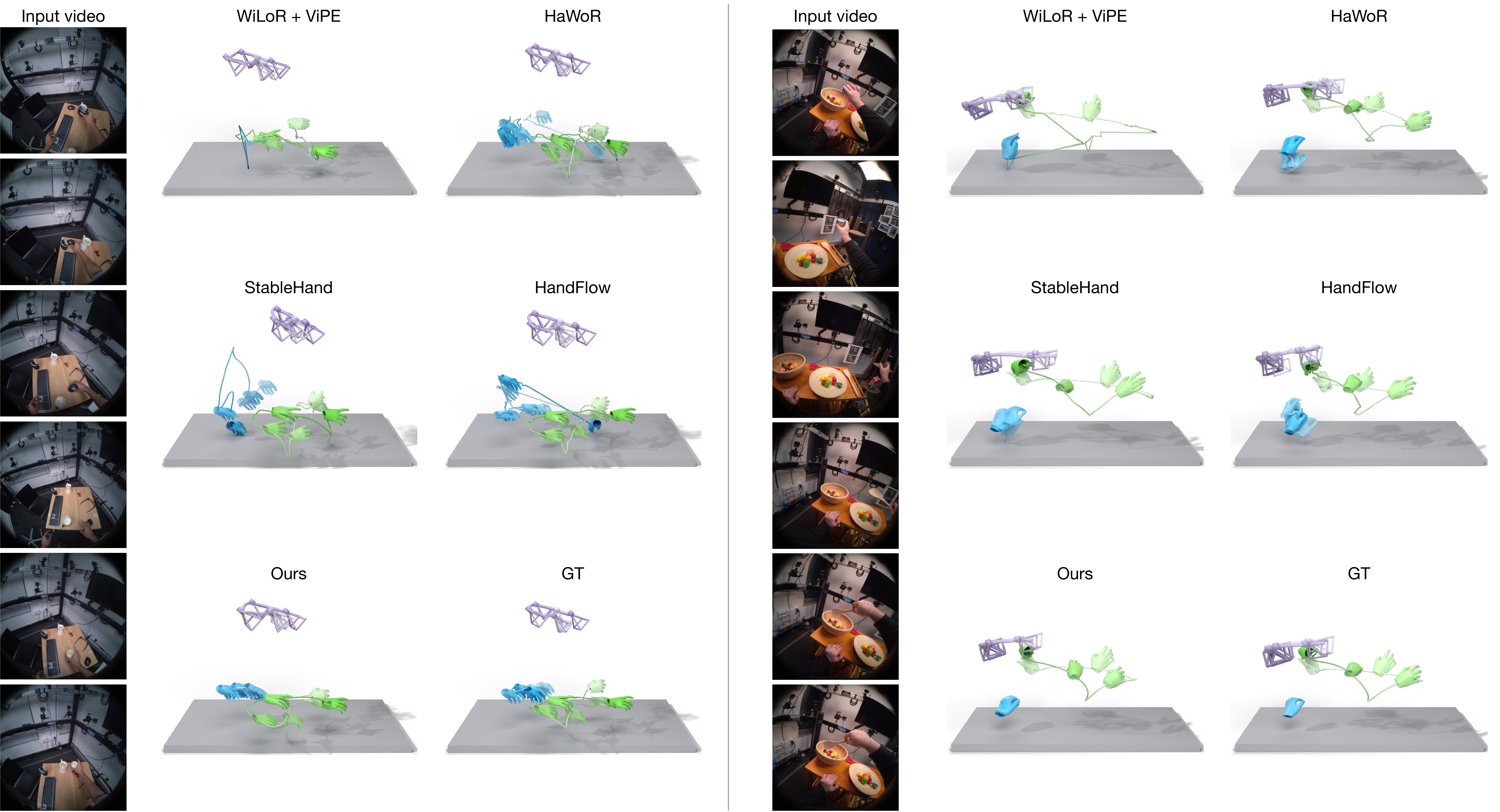}
\vspace{-10pt}
\caption{\textbf{Qualitative comparison on HOT3D.}
Two selected sequences show six input frames and reconstructions from representative methods and ground truth.
For each method, one $\mathrm{SE}(3)$ transform fitted jointly to both hands is also applied to its camera poses; no scale correction is used. Methods share the same view and crop within each sequence.
Green and blue denote the right and left hands; purple frustums indicate camera poses.}
\label{fig:qual_hot3d}
\end{figure}

\customparagraph{Results}
DewHand achieves the best world-space accuracy on both datasets (Table~\ref{tab:main}). Against
StableHand with predicted shape, it reduces $\mathrm{SE}(3)$-aligned WA-MPJPE from $43.74$ to $28.56$\,mm
on the HOT3D StableHand split and from $78.63$ to $36.35$\,mm on ARCTIC, and also improves PA-MPJPE.
With the same privileged ground-truth shape, ours leads in every metric on ARCTIC and trails StableHand
on HOT3D only in PA-MPJPE, MRRPE-A, and (by $0.3$\,mm) $\mathrm{Sim}(3)$-aligned wWA, so the
world-space gains do not come from hand shape. The small gap between our $\mathrm{Sim}(3)$-
and $\mathrm{SE}(3)$-aligned errors indicates that metric scale is retained; ours also has the lowest
AccEr, and the lowest MRRPE-A among predicted-shape methods.
Fig.~\ref{fig:qual_hot3d} illustrates this: DewHand's hand trajectories remain close to ground truth,
whereas several baselines show greater dispersion in the left sequence or displace the nearly
stationary left hand in the right. Its margin persists when more than half of the frames lack a hand
detection ($71.0$ vs.\ $134.3$\,mm for the best baseline; Appendix~\ref{app:missing}), and it runs
$2.6$--$3.7\times$ faster than the SLAM-based pipelines (Appendix~\ref{app:runtime}).

\begin{table}[t]\centering
\caption{\textbf{Camera motion on HOT3D.} $\delta{<}\kappa$ is the percentage of clips whose
trajectory scale is within a factor $\kappa$ of the ground truth. ATE is reported without alignment
(raw) and after $\mathrm{SE}(3)$ or $\mathrm{Sim}(3)$ alignment; ATE/RPE in mm, rot in degrees.
\textbf{Best} in bold, \underline{second best} underlined.}
\label{tab:main_cam}
\small
\setlength{\tabcolsep}{4pt}
\begin{tabular}{lrrrrrrr}
\toprule
Method & $\delta{<}1.10$ $\uparrow$ & $\delta{<}1.25$ $\uparrow$ & ATE$_\mathrm{raw}$ $\downarrow$ & ATE$_\mathrm{SE3}$ $\downarrow$ & ATE$_\mathrm{Sim3}$ $\downarrow$ & rot $\downarrow$ & RPE$_\mathrm{t}$@1 $\downarrow$ \\
\midrule
HaWoR          & \underline{26.1\%} & 42.2\% & 142.6 & 71.2 & 36.2 & 8.27 & \underline{7.61} \\
ViPE           & 24.6\% & \underline{53.6\%} & 96.4  & 45.6 & \underline{24.9} & \underline{4.61} & 10.21 \\
VGGT-$\Omega$  & 20.6\% & 48.8\% & \underline{77.0}  & \underline{38.8} & 27.7 & 6.00 & 12.78 \\
\textbf{Ours}  & \textbf{66.2\%} & \textbf{91.9\%} & \textbf{31.5} & \textbf{16.7} & \textbf{13.1} & \textbf{1.73} & \textbf{3.79} \\
\bottomrule
\end{tabular}
\end{table}

\subsection{Camera Motion}

\customparagraph{Evaluation}
On HOT3D, we report camera scale accuracy, raw and aligned ATE, mean rotation
error, and 1-frame RPE (Table~\ref{tab:main_cam}); all trajectories are relative to frame~0.
\customparagraph{Baselines}
We compare against VGGT-$\Omega$, the frozen geometry backbone we condition on, together
with the two SLAM systems used to lift the hand baselines: HaWoR's~\citep{hawor} internal
DROID-SLAM~\citep{droidslam} and ViPE~\citep{vipe}.
\customparagraph{Results}
Despite conditioning on frozen VGGT-$\Omega$ features, DewHand improves camera scale and
trajectory accuracy over the backbone (Table~\ref{tab:main_cam}). Gains after similarity
alignment indicate improvements beyond scale correction.

\begin{table}[t]\centering
\caption{\textbf{Zero-shot generalization to HoloAssist, H2O, and TACO.}
All methods are scored on common frames. Camera-space methods are lifted to world coordinates with
ViPE run on each 150-frame evaluation window with ground-truth intrinsics; DewHand uses no camera
intrinsics. $\dagger$ marks target-domain training; \textbf{best} and \underline{second best} are
marked within each dataset.}
\label{tab:zeroshot}
\label{tab:holoassist}
\label{tab:h2o}
\label{tab:taco}
\small
\setlength{\tabcolsep}{2pt}
\renewcommand{\arraystretch}{1.12}
\begin{tabular*}{\linewidth}{@{\extracolsep{\fill}}llrrrrrrrrr@{}}
\toprule
 & \multirow{2}{*}{Method} & \multirow{2}{*}{PA $\downarrow$} & \multicolumn{2}{c}{W-MPJPE $\downarrow$} & \multicolumn{2}{c}{WA-MPJPE $\downarrow$} & \multicolumn{2}{c}{wWA $\downarrow$} & \multirow{2}{*}{AccEr $\downarrow$} & \multirow{2}{*}{MRRPE-A $\downarrow$} \\
\cmidrule(lr){4-5}\cmidrule(lr){6-7}\cmidrule(lr){8-9}
 & & & $\mathrm{Sim}(3)$ & $\mathrm{SE}(3)$ & $\mathrm{Sim}(3)$ & $\mathrm{SE}(3)$ & $\mathrm{Sim}(3)$ & $\mathrm{SE}(3)$ & & \\
\midrule
\multirow{7}{*}{\rotatebox[origin=c]{90}{\textbf{HoloAssist}}} & HaWoR & 13.47 & \underline{128.07} & \underline{148.96} & \underline{51.59} & \underline{77.36} & \underline{38.90} & \underline{67.87} & 24.91 & 112.25 \\
 & HandFlow & \underline{12.53} & 142.57 & 233.89 & 55.39 & 161.42 & 43.36 & 153.23 & 29.91 & 109.69 \\
 & ReViV$^{\dagger}$ & 14.17 & 224.11 & 253.67 & 63.02 & 125.99 & 46.82 & 113.63 & \underline{23.71} & 136.57 \\
 & WiLoR\,$+$\,ViPE & 12.69 & 167.78 & 263.88 & 56.24 & 179.21 & 43.73 & 174.46 & 30.31 & 139.42 \\
 & EgoForce\,$+$\,ViPE & 13.78 & 172.61 & 279.13 & 68.35 & 184.37 & 52.17 & 176.22 & 46.20 & 156.19 \\
 & HOPformer\,$+$\,ViPE & 16.54 & 150.57 & 232.90 & 59.40 & 153.31 & 40.65 & 137.55 & 31.86 & \underline{91.82} \\
 & \textbf{Ours} & \textbf{11.96} & \textbf{105.00} & \textbf{112.86} & \textbf{40.89} & \textbf{52.67} & \textbf{29.42} & \textbf{42.25} & \textbf{7.33} & \textbf{84.55} \\
\midrule
\multirow{7}{*}{\rotatebox[origin=c]{90}{\textbf{H2O}}} & HaWoR & 7.22 & 62.61 & 62.86 & 22.81 & 35.57 & 14.46 & 31.63 & 13.35 & 34.52 \\
 & HandFlow & 7.32 & \underline{32.76} & 62.67 & 17.26 & 25.13 & 11.51 & 20.87 & 8.72 & 36.59 \\
 & ReViV$^{\dagger}$ & 9.44 & 56.76 & \underline{55.46} & 23.72 & 27.45 & 12.93 & 18.96 & \underline{3.05} & 45.28 \\
 & WiLoR\,$+$\,ViPE & \textbf{5.68} & 35.93 & 64.65 & \underline{14.80} & \underline{22.04} & \textbf{7.94} & \underline{16.26} & 7.50 & \underline{23.76} \\
 & EgoForce\,$+$\,ViPE$^{\dagger}$ & \underline{6.08} & \textbf{27.68} & 58.75 & \textbf{14.16} & 22.50 & \underline{9.08} & 18.50 & 7.58 & \textbf{22.76} \\
 & HOPformer\,$+$\,ViPE & 9.67 & 51.92 & 81.68 & 25.22 & 32.87 & 13.59 & 23.02 & 12.18 & 48.49 \\
 & \textbf{Ours} & 6.80 & 40.88 & \textbf{40.77} & 19.54 & \textbf{20.68} & 12.10 & \textbf{15.05} & \textbf{2.60} & 46.17 \\
\midrule
\multirow{7}{*}{\rotatebox[origin=c]{90}{\textbf{TACO}}} & HaWoR & 9.77 & 80.84 & 84.61 & 32.34 & 43.63 & 23.95 & 38.01 & 15.47 & 57.17 \\
 & HandFlow & 10.49 & \textbf{55.51} & \textbf{56.76} & 27.91 & 36.05 & 19.89 & 29.80 & 9.82 & 72.61 \\
 & ReViV$^{\dagger}$ & 11.41 & 88.09 & 90.67 & 40.07 & 47.77 & 27.86 & 39.72 & \underline{5.63} & \textbf{54.28} \\
 & WiLoR\,$+$\,ViPE & \textbf{8.51} & 56.98 & 60.94 & \underline{27.38} & \underline{33.45} & \underline{19.71} & 29.29 & 12.37 & \underline{55.82} \\
 & EgoForce\,$+$\,ViPE & 10.13 & 66.52 & 71.55 & 33.91 & 43.06 & 25.04 & 38.17 & 17.28 & 75.41 \\
 & HOPformer\,$+$\,ViPE & 12.70 & 74.58 & 75.48 & 38.50 & 41.73 & 22.97 & \underline{28.58} & 17.08 & 90.59 \\
 & \textbf{Ours} & \underline{9.26} & \underline{56.91} & \underline{58.99} & \textbf{26.76} & \textbf{31.29} & \textbf{18.58} & \textbf{24.15} & \textbf{3.35} & 60.95 \\
\bottomrule
\end{tabular*}

\end{table}

\subsection{Cross-Device Zero-Shot Generalization}
We evaluate the HOT3D-trained model without fine-tuning on HoloAssist~\citep{holoassist},
testing transfer from Aria fisheye video to HoloLens~2 perspective video. HoloAssist provides
camera poses and articulated hand tracks, enabling world-space evaluation across capture devices.
On randomly sampled validation windows, DewHand achieves the lowest errors across all reported
metrics, including PA-MPJPE, world-space placement, acceleration, and inter-hand placement
(Table~\ref{tab:holoassist}). Camera-space baselines are lifted with per-window ViPE~\citep{vipe} rather than ground-truth cameras; target-domain-trained references are marked in the table.

We further evaluate zero-shot transfer on the held-out subject of the H2O~\citep{h2o} pose
benchmark and on TACO~\citep{taco}, lifting the camera-space methods with per-window ViPE
trajectories. On both datasets, DewHand achieves the lowest $\mathrm{SE}(3)$-aligned joint and
wrist-level WA-MPJPE, as well as AccEr (Table~\ref{tab:zeroshot}).
Other metrics, particularly local pose and inter-hand placement, show mixed results.

\subsection{Ablation Study}

We ablate camera supervision, visual conditioning, and the training objective on HOT3D.

\customparagraph{Joint Camera Estimation {\mdseries (Table~\ref{tab:abl64_all}(a))}}
Removing camera supervision increases $\mathrm{SE}(3)$-aligned WA-MPJPE from $35.94$ to
$42.36$\,mm while leaving PA-MPJPE nearly unchanged. Camera supervision therefore primarily
improves global hand placement rather than local articulation.

\customparagraph{Conditioning Features {\mdseries (Table~\ref{tab:abl64_all}(b))}}
Replacing VGGT-$\Omega$ with DINOv2~\citep{dinov2} raises $\mathrm{SE}(3)$-aligned WA-MPJPE from $35.94$ to $49.15$\,mm
and generated-camera ATE from $31.5$ to $88.9$\,mm, supporting the value of 3D geometric features.
DA3~\citep{depthanything3} instead yields comparable hand and camera accuracy, despite its backbone camera ATE being
$113.3$ versus $77.0$\,mm for VGGT-$\Omega$. This suggests that the camera trajectory is generated
rather than copied from the backbone estimate.
Removing hand-specific features substantially worsens local pose while leaving camera accuracy
stable, indicating that the two condition streams play complementary roles.

\customparagraph{Training Objective {\mdseries (Table~\ref{tab:abl64_all}(c))}}
Compared with L2 regression, flow matching improves local pose (PA-MPJPE: $7.04$ to $6.76$\,mm),
$\mathrm{SE}(3)$-aligned WA-MPJPE ($45.13$ to $35.94$\,mm), and temporal accuracy (AccEr: $3.34$ to
$1.89$\,$\mathrm{m/s^2}$), while camera accuracy remains comparable. The gain does not come from the
architecture, which the regressor shares, and inference does not reduce to a single prediction:
one Euler step, which returns the average of the network's direct estimates, is $3.4$\,mm worse and nearly doubles AccEr
(Appendix~\ref{app:sampling_settings}). Flow samples also spread more where the hand is undetected
($8.9$ vs.\ $23.8$\,mm), and the spread tracks the error (Appendix~\ref{app:complement}).

\begin{table}[t]\centering
\caption{\textbf{Ablations on HOT3D.} (a) Camera supervision, (b) visual conditioning,
and (c) training objective. All variants use mask tokens and the same 64-epoch training recipe; each changes only the named component.
\textbf{Best} in bold and \underline{second best} underlined within each block.
Metrics follow Tables~\ref{tab:main} and~\ref{tab:main_cam}; scale accuracy is reported as a percentage.}
\label{tab:abl64_all}
\small
\setlength{\tabcolsep}{2pt}
\renewcommand{\arraystretch}{1.08}
\begin{tabular*}{\linewidth}{@{\extracolsep{\fill}}lrrrrrrrr@{}}
\toprule
 & \multicolumn{4}{c}{Hands} & \multicolumn{4}{c}{Camera} \\
\cmidrule(lr){2-5}\cmidrule(lr){6-9}
Variant & PA $\downarrow$ & WA$_{\mathrm{SE}(3)}$ $\downarrow$ & wWA$_{\mathrm{SE}(3)}$ $\downarrow$ & AccEr $\downarrow$ & $\delta\!<\!1.10$ $\uparrow$ & $\delta\!<\!1.25$ $\uparrow$ & ATE$_\mathrm{raw}$ $\downarrow$ & rot $\downarrow$ \\
\midrule
\multicolumn{9}{@{}l@{}}{\colorbox{black!8}{\strut\textbf{(a) Camera supervision}}} \\
\textbf{Full} & \textbf{6.76} & \textbf{35.94} & \textbf{29.49} & 1.89 & \textbf{66.2} & \textbf{91.9} & \textbf{31.5} & \textbf{1.73} \\
w/o camera supervision & 6.86 & 42.36 & 34.63 & \textbf{1.80} & -- & -- & -- & -- \\
\midrule
\multicolumn{9}{@{}l@{}}{\colorbox{black!8}{\strut\textbf{(b) Conditioning features}}} \\
\textbf{\makebox[4.1em][l]{HaMeR} + VGGT-$\Omega$} & \underline{6.76} & \textbf{35.94} & \textbf{29.49} & \underline{1.89} & \textbf{66.2} & \textbf{91.9} & \textbf{31.5} & \underline{1.73} \\
\makebox[4.1em][l]{HaMeR} + DA3 & 6.77 & \underline{36.40} & \underline{29.86} & \textbf{1.79} & \underline{64.2} & \underline{90.6} & \textbf{31.5} & \textbf{1.60} \\
\makebox[4.1em][l]{HaMeR} + DINOv2 & \textbf{6.71} & 49.15 & 43.20 & 1.97 & 34.0 & 63.0 & 88.9 & 6.12 \\
\makebox[4.1em][l]{$\varnothing$} + VGGT-$\Omega$ & 10.72 & 46.57 & 35.07 & 2.04 & 64.0 & 90.3 & 32.8 & 1.83 \\
\midrule
\multicolumn{9}{@{}l@{}}{\colorbox{black!8}{\strut\textbf{(c) Training objective}}} \\
L2 regression & 7.04 & 45.13 & 38.13 & 3.34 & 62.0 & 89.8 & \textbf{31.4} & 1.83 \\
Flow matching & \textbf{6.76} & \textbf{35.94} & \textbf{29.49} & \textbf{1.89} & \textbf{66.2} & \textbf{91.9} & 31.5 & \textbf{1.73} \\
\bottomrule
\end{tabular*}
\end{table}

\section{Conclusion}
We addressed metric world-space reconstruction of two hands and the camera from monocular
egocentric video. DewHand jointly generates global hand and camera motion, local articulation,
and hand shape in a shared world frame, avoiding the composition of separately estimated
trajectories. Geometric and hand-specific features provide complementary cues for global placement
and articulation, while flow matching models motion under incomplete observations. On HOT3D and
ARCTIC, DewHand achieves the lowest rigid-aligned world-space hand error. The small gap between
rigid- and similarity-aligned hand errors further supports metric-scale recovery without scale
correction at evaluation. On HOT3D, it also improves metric camera scale recovery over the compared
SLAM systems. Camera supervision primarily
improves global hand placement, and flow matching yields more accurate and temporally consistent
motion than direct regression. Zero-shot results show the lowest rigid-aligned WA-MPJPE on HoloAssist,
H2O, and TACO, although other metrics do not always lead. Future work could
extend the model beyond fixed offline windows and incorporate manipulated objects and hand-object
contact.

\section*{Acknowledgment}

This work is supported by JST, CRONOS, Japan Grant Number 	JPMJCS24K6, JST FOREST, Grant Number JPMJFR252X, JST BOOST, Grant Number JPMJBY24C5, JST Moonshot JPMJMS263D, JST CREST, Japan, Grant Number JPMJCR26X4 and JPMJCR2654, JST K Program Grant Number JPMJKP25V2.

\bibliographystyle{iclr2027_conference}
\bibliography{iclr2027_conference}

\clearpage

\appendix

\section{Implementation Details}
\label{app:implementation}
During training, we randomly mask each valid hand observation with probability $0.10$, one
contiguous span of 4 to 16 frames with probability $0.35$ per hand and clip, and a complete hand
track with probability $0.05$. The model width is $D=512$, with six attention layers and 16
heads per layer.

\section{Evaluation protocol}
\label{app:protocol}
\subsection{Hand metrics}
Let $\hat{\mathbf{j}}^h_{t,k}$ and $\mathbf{j}^h_{t,k}$ be the predicted and ground-truth positions
of joint $k\in\{0,\dots,20\}$ ($k=0$ is the wrist) of hand $h$ at frame $t$, and let $\mathcal V$ be
the ground-truth-valid frames of a hand-clip. The MPJPE variants average the error after aligning
the prediction with a transform $\mathcal T^\star$,
\begin{equation}
\mathrm{E}=\frac{1}{21\,|\mathcal V|}\sum_{t\in\mathcal V}\sum_{k=0}^{20}
\big\|\mathcal T^\star(\hat{\mathbf{j}}^h_{t,k})-\mathbf{j}^h_{t,k}\big\|_2,
\qquad
\mathcal T^\star=\operatorname*{arg\,min}_{\mathcal T\in G}\sum_{(t,k)\in\mathcal S}
\big\|\mathcal T(\hat{\mathbf{j}}^h_{t,k})-\mathbf{j}^h_{t,k}\big\|_2^2,
\end{equation}
and differ in the transformation group $G$ and the frames $\mathcal S$ used to fit it:
\begin{itemize}[topsep=2pt,itemsep=1pt,parsep=0pt,leftmargin=1.5em]
\item \textbf{PA-MPJPE}: $G=\mathrm{Sim}(3)$, with a separate $\mathcal T^\star_t$ fitted at each frame
($\mathcal S=\{t\}\times\{0,\dots,20\}$); measures local articulation.
\item \textbf{W-MPJPE}: $G=\mathrm{Sim}(3)$ or $\mathrm{SE}(3)$, fitted on the first two frames and applied
to the whole clip, so later drift remains; this follows the evaluation code of HaWoR~\citep{hawor}
and HandFlow~\citep{handflow}.
\item \textbf{WA-MPJPE}: $G=\mathrm{Sim}(3)$ or $\mathrm{SE}(3)$, fitted on all frames; measures global
placement. $\mathrm{SE}(3)$ keeps the predicted scale, so scale errors remain.
\item \textbf{wWA}: WA-MPJPE computed on the wrist ($k=0$) only. For DewHand, which predicts the wrist directly,
wWA is nearly unaffected by the shape used for posing (at most $0.03$\,mm between the Pred.- and GT-shape
rows of Table~\ref{tab:main}); StableHand instead takes the shape as a network input, so its whole
predicted trajectory, including the wrist, changes with the input shape.
\item \textbf{AccEr}: error of the joint acceleration, with no alignment ($\mathrm{m/s^2}$),
\[
\frac{1}{21|\mathcal V'|}\sum_{t\in\mathcal V'}\sum_{k=0}^{20}
\frac{\|\Delta^2\hat{\mathbf{j}}^h_{t,k}-\Delta^2\mathbf{j}^h_{t,k}\|_2}{\Delta t^2},
\qquad \Delta^2\mathbf{j}_{t}=\mathbf{j}_{t+1}-2\mathbf{j}_{t}+\mathbf{j}_{t-1},
\]
where $\Delta t=1/30$\,s and $\mathcal V'$ are the frames whose neighbors are valid.
\item \textbf{MRRPE-A}: error of the inter-wrist displacement over the frames $\mathcal B$ where both
hands are valid,
\[
\frac{1}{|\mathcal B|}\sum_{t\in\mathcal B}
\big\|\mathbf{R}^\star(\hat{\mathbf{j}}^{R}_{t,0}-\hat{\mathbf{j}}^{L}_{t,0})-(\mathbf{j}^{R}_{t,0}-\mathbf{j}^{L}_{t,0})\big\|_2,
\]
where $\mathbf{R}^\star$ is the rotation that best aligns
the joints of both hands jointly; the shared translation cancels and no scale is applied.
\end{itemize}

\subsection{Camera metrics}
All camera trajectories are expressed relative to frame~0. ATE is the root-mean-square
distance between predicted and ground-truth camera centers, measured directly (raw) or
after one clip-level rigid ($\mathrm{SE}(3)$) or similarity ($\mathrm{Sim}(3)$) alignment.
Scale accuracy $\delta<\kappa$ is the percentage of clips satisfying
$\max(s,1/s)<\kappa$, where $s$ is the predicted-to-ground-truth trajectory scale ratio.
The rotation error is the mean geodesic distance between corresponding camera rotations;
RPE$_{\mathrm t}$@1 is the root-mean-square error of one-frame relative camera translations.

\subsection{Evaluation sets}
\label{app:holoassist}\label{app:evalsets}
Table~\ref{tab:evalsets} lists the evaluation sets; all windows have $150$ frames at 30\,fps.
Every method is scored by us with the same metric definitions on the same ground-truth-annotated frames (HOPformer
on ARCTIC from its released predictions),
without visibility gating. Missing per-frame predictions of baselines are linearly interpolated,
and a clip without output from some method is dropped from the common set. MRRPE-A uses the clips
in which both hands are annotated and predicted.

\begin{table}[H]\centering
\caption{\textbf{Evaluation sets.} Camera-space baselines are lifted with ViPE, run on the full
sequence for HOT3D and ARCTIC and per window (with ground-truth intrinsics) for the zero-shot datasets.}
\label{tab:evalsets}
\small
\setlength{\tabcolsep}{6pt}
\begin{tabular}{llrr}
\toprule
Dataset & Split & Windows & Hand-clips \\
\midrule
HOT3D & HaWoR split & 606 & 1{,}205 \\
HOT3D & StableHand split & 244 & 485 \\
ARCTIC & validation (s05) & 155 & 310 \\
HoloAssist & validation & 410 & 597 \\
H2O & \texttt{pose\_test} (subject 4) & 180 & 321 \\
TACO & -- & 133 & 262 \\
\bottomrule
\end{tabular}
\end{table}

HandFlow handles only the right hand, so we obtain left-hand predictions by mirroring the video.
We use StableHand's released ``enhanced'' checkpoint and compare only against StableHand on its
split because HaWoR was trained on those sequences. The StableHand-split result uses a separate
DewHand model, trained on HOT3D training windows after removing
every window that overlaps the 244 StableHand test clips; other segments of the same sequences remain
in training, as in StableHand's own protocol. For HoloAssist, we map the 26-joint HoloLens
tracks to 21 joints. ReViV's two-second predictions are rigidly stitched into 150-frame sequences.

\section{Additional results}
\subsection{Inference steps}
\label{app:sampling_settings}
For velocity averaging, we initialize five trajectories from independent Gaussian noise and average
their predicted flow fields at each Euler step.
\begin{table}[H]\centering
\caption{\textbf{Effect of Euler steps at inference on HOT3D}, for our full model with five-sample velocity
averaging. Metrics as in Tables~\ref{tab:main} and~\ref{tab:main_cam}. \textbf{Best} in bold,
\underline{second best} underlined.}
\label{tab:steps}
\small
\setlength{\tabcolsep}{5pt}
\begin{tabular}{crrrrrrr}
\toprule
 & & & & \multicolumn{4}{c}{Camera} \\
\cmidrule(lr){5-8}
Steps & PA $\downarrow$ & WA$_{\mathrm{SE}(3)}$ $\downarrow$ & AccEr $\downarrow$ & $\delta\!<\!1.10$ $\uparrow$ & $\delta\!<\!1.25$ $\uparrow$ & ATE$_\mathrm{raw}$ $\downarrow$ & rot $\downarrow$ \\
\midrule
1  & 6.84 & 39.35 & 3.64 & 62.2 & 90.8 & 33.1 & 1.92 \\
\textbf{2} & \underline{6.76} & \textbf{35.94} & 1.89 & \textbf{66.2} & \textbf{91.9} & \textbf{31.5} & \textbf{1.73} \\
5  & \textbf{6.75} & \underline{36.06} & 1.65 & \underline{65.0} & \underline{91.7} & \underline{31.7} & \underline{1.74} \\
10 & \underline{6.76} & 36.50 & \textbf{1.63} & 64.0 & \underline{91.7} & 31.9 & 1.75 \\
20 & 6.79 & 37.02 & \underline{1.64} & 62.4 & \textbf{91.9} & 32.2 & 1.77 \\
\bottomrule
\end{tabular}
\end{table}

Table~\ref{tab:steps} compares the number of Euler steps. Two steps achieve the lowest world-aligned
hand error and camera trajectory error, while additional steps reduce AccEr but do not
improve world-space placement. We therefore use two steps by default.

\subsection{Sampling with missing observations}
\label{app:complement}
Table~\ref{tab:complement} compares frames with and without WiLoR hand detections. Sample spread
grows without detections and tracks the error (Spearman $\rho=0.58$ over all frames), while our
method has lower WA-MPJPE and AccEr than L2 regression in both groups.
\begin{table}[H]\centering
\caption{\textbf{Sampling with incomplete hand observations on HOT3D.}
Frames are split by whether WiLoR detects the hand. Sample spread measures how much five
independent samples of our model disagree: the per-frame standard deviation of their joint
positions, averaged over joints (no ground truth used). WA-MPJPE is averaged over the frames of
each group. ``Mean of 5'' averages the five predictions before scoring; the AccEr of Ours is
computed per sample and averaged. \emph{Best of 5} picks the sample closest
to ground truth and is not realizable.}
\label{tab:complement}
\small
\setlength{\tabcolsep}{8pt}
\begin{tabular}{lrr}
\toprule
 & Detected & Not detected \\
\midrule
Frames (\%) & 74 & 26 \\
\midrule
\multicolumn{3}{@{}l}{\textit{Sample spread (mm)}} \\
\quad Ours & 8.9 & 23.8 \\
\midrule
\multicolumn{3}{@{}l}{\textit{WA-MPJPE, $\mathrm{SE}(3)$ (mm)}} \\
\quad Ours, mean of 5 & 26.8 & 62.2 \\
\quad Ours, best of 5 & \emph{22.1} & \emph{50.3} \\
\quad L2 regression & 35.3 & 72.0 \\
\midrule
\multicolumn{3}{@{}l}{\textit{AccEr ($\mathrm{m/s^2}$)}} \\
\quad Ours & 2.02 & 2.54 \\
\quad L2 regression & 2.81 & 3.86 \\
\bottomrule
\end{tabular}
\end{table}

\subsection{Detection failures}
\label{app:missing}
Table~\ref{tab:missing} reports world-space hand error by the proportion of frames without a
WiLoR detection. DewHand has the lowest error in every group, including clips with most frames
undetected.
\begin{table}[H]\centering
\caption{\textbf{World-space hand error by detection-failure rate on HOT3D.}
The $606$ clips with valid hand ground truth are grouped by the fraction of annotated frames
without a WiLoR hand detection. We report clip-level WA-MPJPE after rigid $\mathrm{SE}(3)$
alignment (mm) on the same annotated frames for all methods.
$\dagger$: training data overlap with the HOT3D evaluation sequences.}
\label{tab:missing}
\small
\setlength{\tabcolsep}{6pt}
\begin{tabular}{lrrr}
\toprule
Undetected frames & 0\% to 20\% & 20\% to 50\% & $>$50\% \\
\midrule
Clips & 277 & 233 & 96 \\
\midrule
WiLoR\,$+$\,ViPE$^{\dagger}$ & 51.9 & 104.3 & 178.6 \\
HaWoR & 45.9 & 93.1 & 175.5 \\
HandFlow$^{\dagger}$ & 50.6 & 107.0 & 195.7 \\
EgoForce\,$+$\,ViPE$^{\dagger}$ & 51.2 & 90.4 & 134.3 \\
\textbf{Ours} & \textbf{21.6} & \textbf{38.6} & \textbf{71.0} \\
\bottomrule
\end{tabular}
\end{table}

\subsection{Inference Cost}
\label{app:runtime}
Table~\ref{tab:runtime} compares inference time. DewHand is $2.6$--$3.7\times$ faster than the
SLAM-based pipelines because it generates the camera in a few flow steps instead of optimizing it;
of its 12.0\,s, VGGT-$\Omega$ and the flow head take 1.7\,s, and the rest is rectification, hand
detection, and HaMeR feature extraction. StableHand is omitted because its full preprocessing
pipeline is unavailable.
\begin{table}[htbp]
\centering
\caption{\textbf{Inference cost on HOT3D.} Wall-clock time for one 150-frame clip on a single A100 GPU,
including each method's preprocessing (rectification, detection, feature extraction, and SLAM if
used) but not model loading or JPEG decoding. WA-MPJPE ($\mathrm{SE}(3)$, mm) is from Table~\ref{tab:main};
$\dagger$: training data overlap with the HOT3D evaluation sequences.}
\label{tab:runtime}
\small
\setlength{\tabcolsep}{6pt}
\begin{tabular}{lrr}
\toprule
Method & WA-MPJPE $\downarrow$ & Total (s) $\downarrow$ \\
\midrule
WiLoR\,$+$\,ViPE$^{\dagger}$                & 92.06          & 34.4 \\
HaWoR                                        & 84.80          & 44.9 \\
HandFlow$^{\dagger}$                         & 95.24          & 31.4 \\
\midrule
\textbf{Ours}                               & \textbf{35.94} & \textbf{12.0} \\
\bottomrule
\end{tabular}
\end{table}

\section{Additional qualitative comparisons}
\label{app:qualitative}
\begin{figure}[p]
\centering
\includegraphics[width=\linewidth,height=.43\textheight,keepaspectratio]{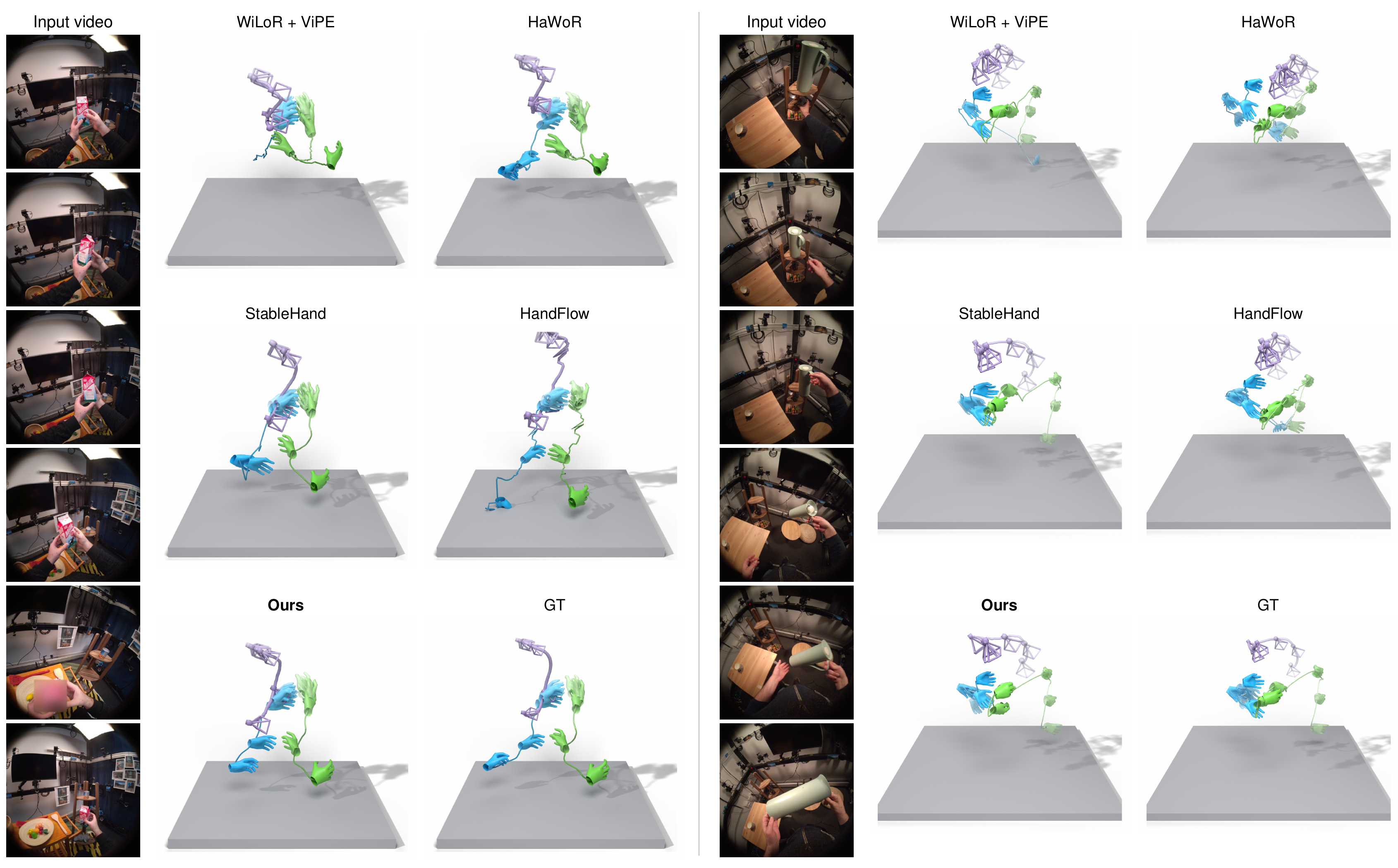}\par
\vspace{10pt}
\includegraphics[width=\linewidth,height=.43\textheight,keepaspectratio]{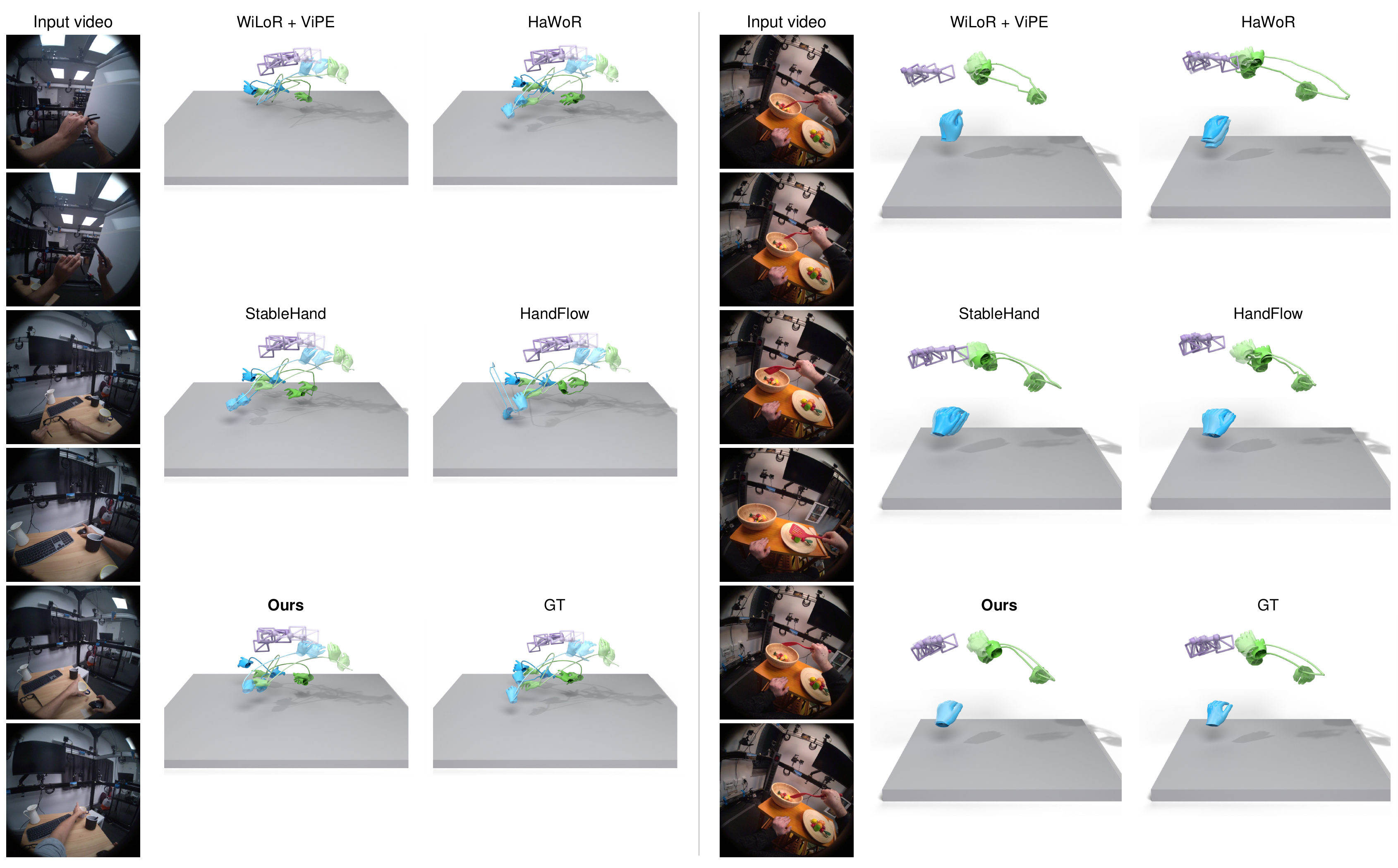}
\vspace{-5pt}
\caption{\textbf{Qualitative comparison on HOT3D.} Four sequences are shown, two per row.}
\label{fig:qual_hot3d_full}
\end{figure}

\begin{figure}[p]
\centering
\includegraphics[width=\linewidth,height=.3\textheight,keepaspectratio]{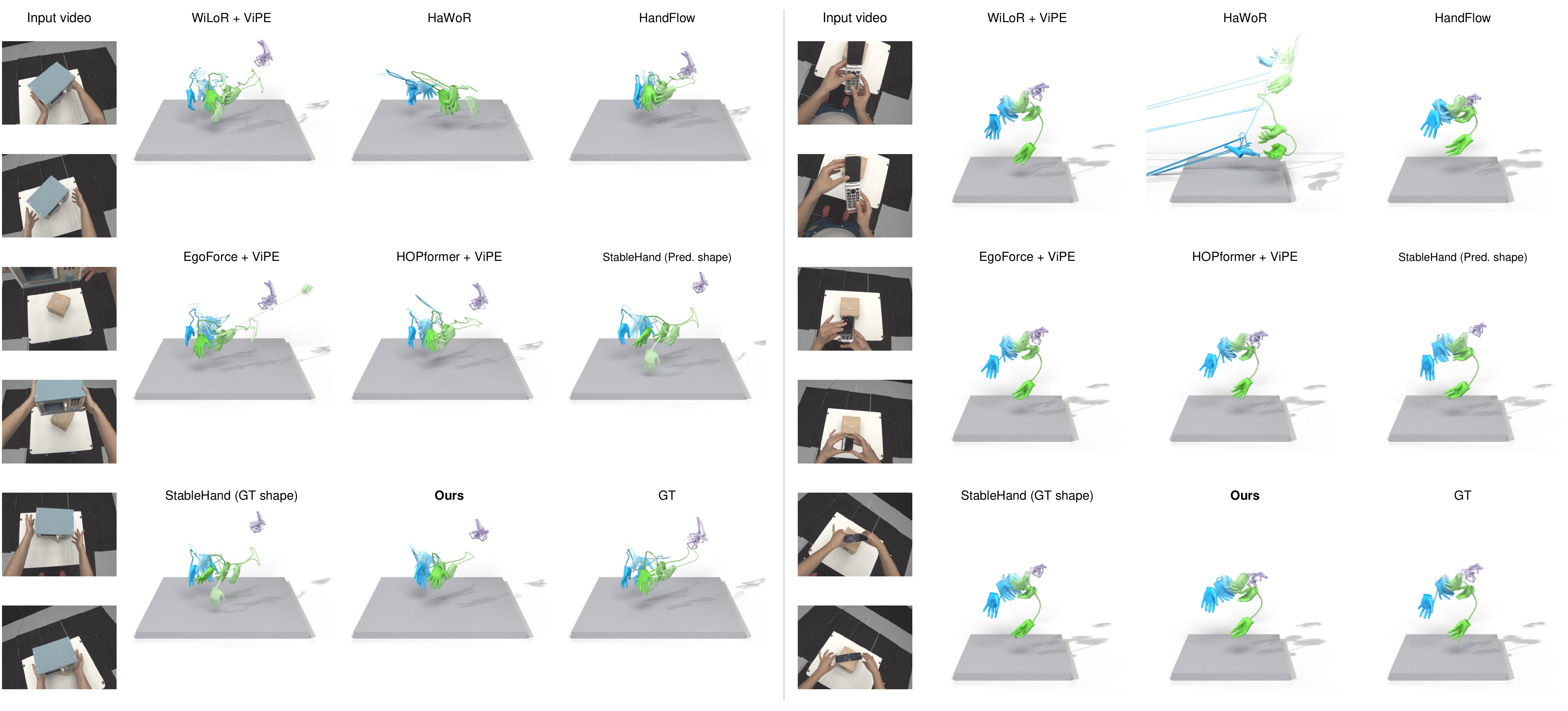}\par
\vspace{10pt}
\includegraphics[width=\linewidth,height=.3\textheight,keepaspectratio]{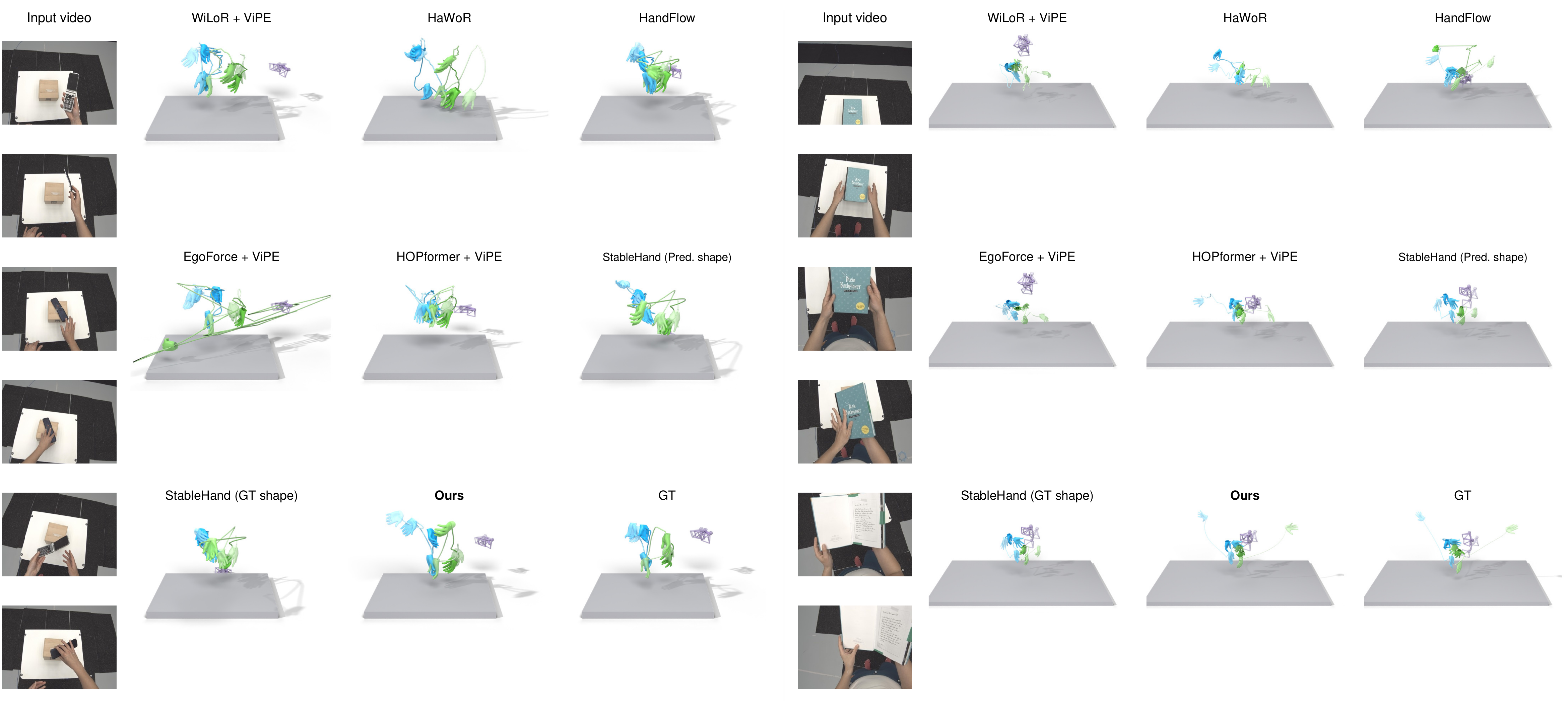}\par
\vspace{10pt}
\includegraphics[width=\linewidth,height=.3\textheight,keepaspectratio]{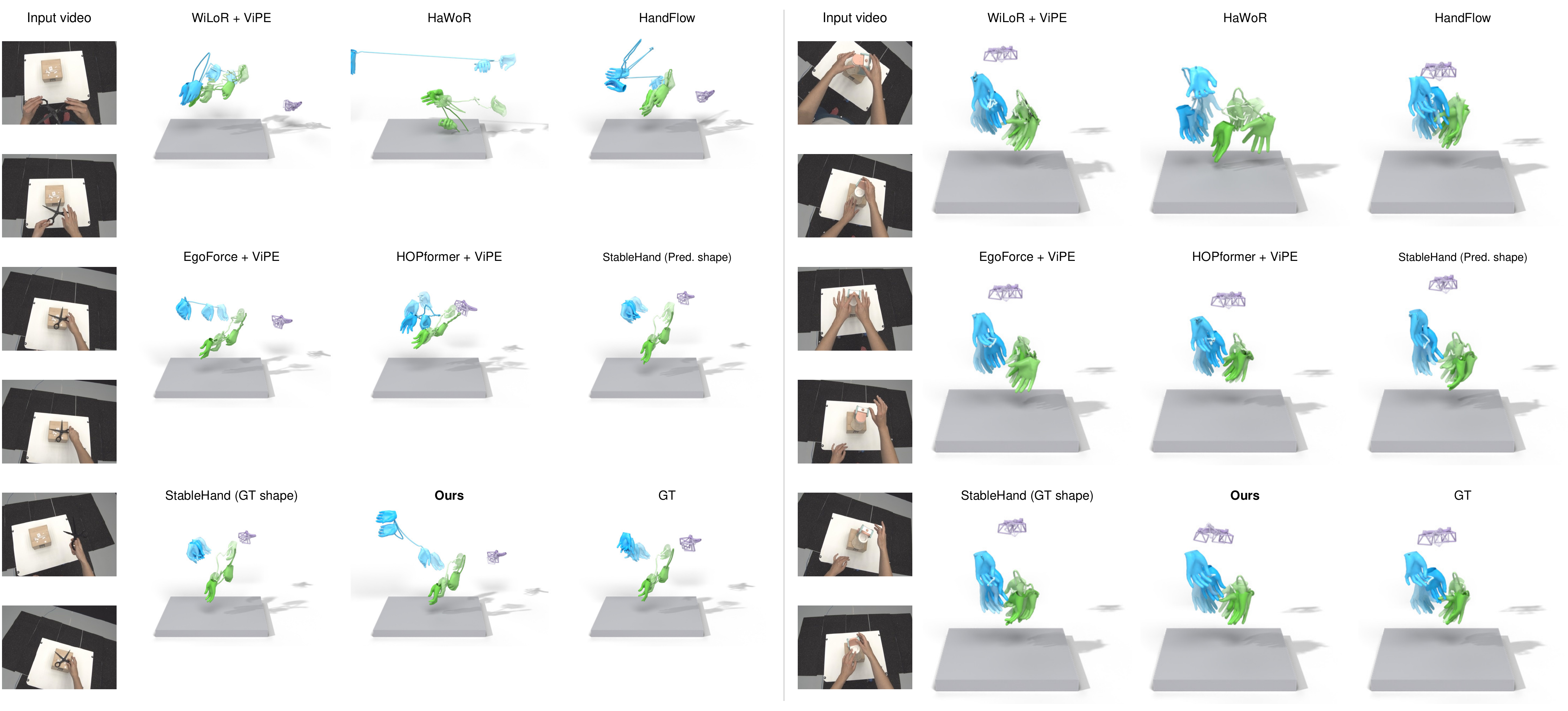}
\vspace{-5pt}
\caption{\textbf{Qualitative comparison on ARCTIC.} Six sequences are shown, two per row.}
\label{fig:qual_arctic}
\end{figure}

\begin{figure}[p]
\centering
\includegraphics[width=\linewidth,height=.43\textheight,keepaspectratio]{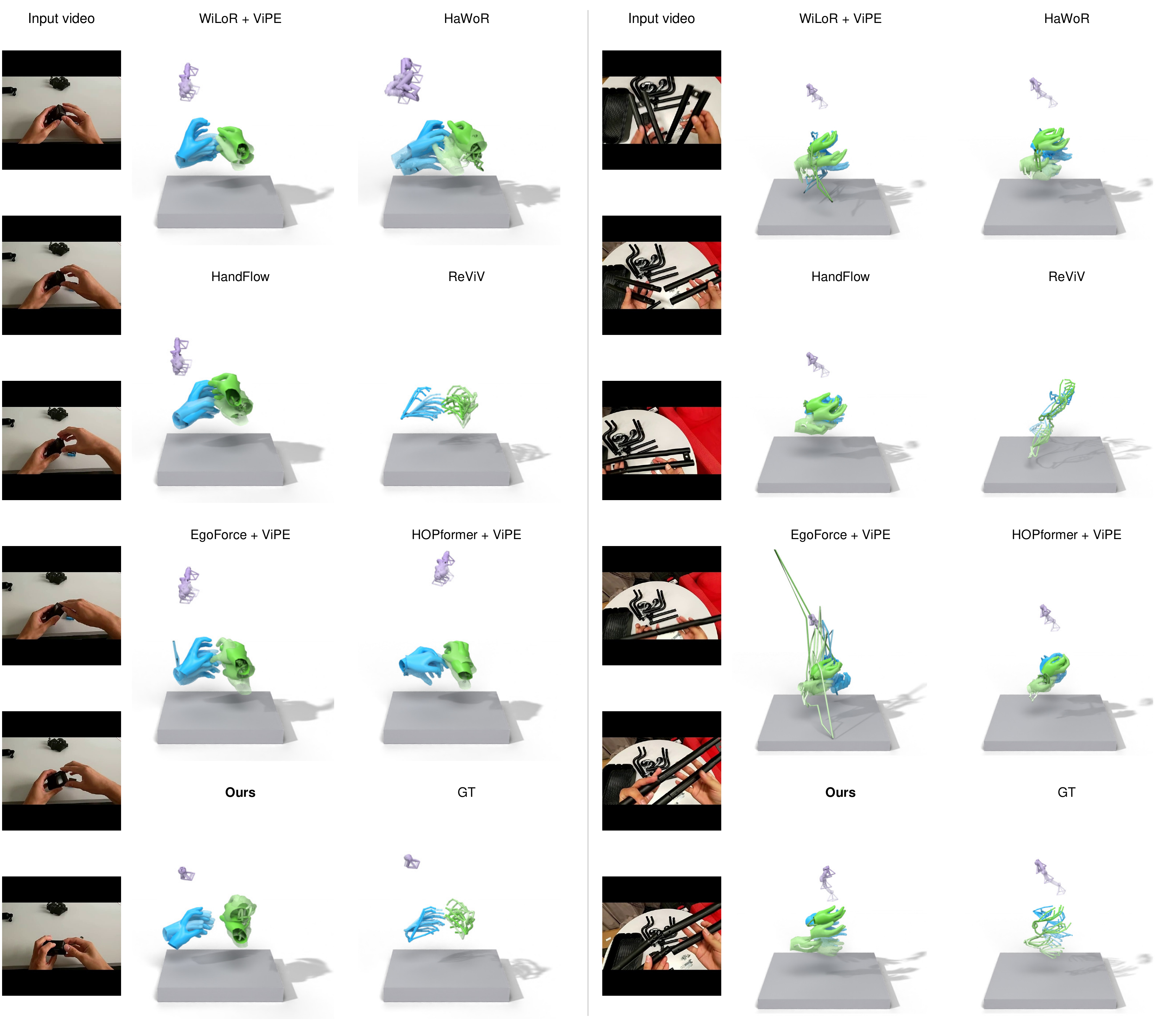}\par
\vspace{10pt}
\includegraphics[width=\linewidth,height=.43\textheight,keepaspectratio]{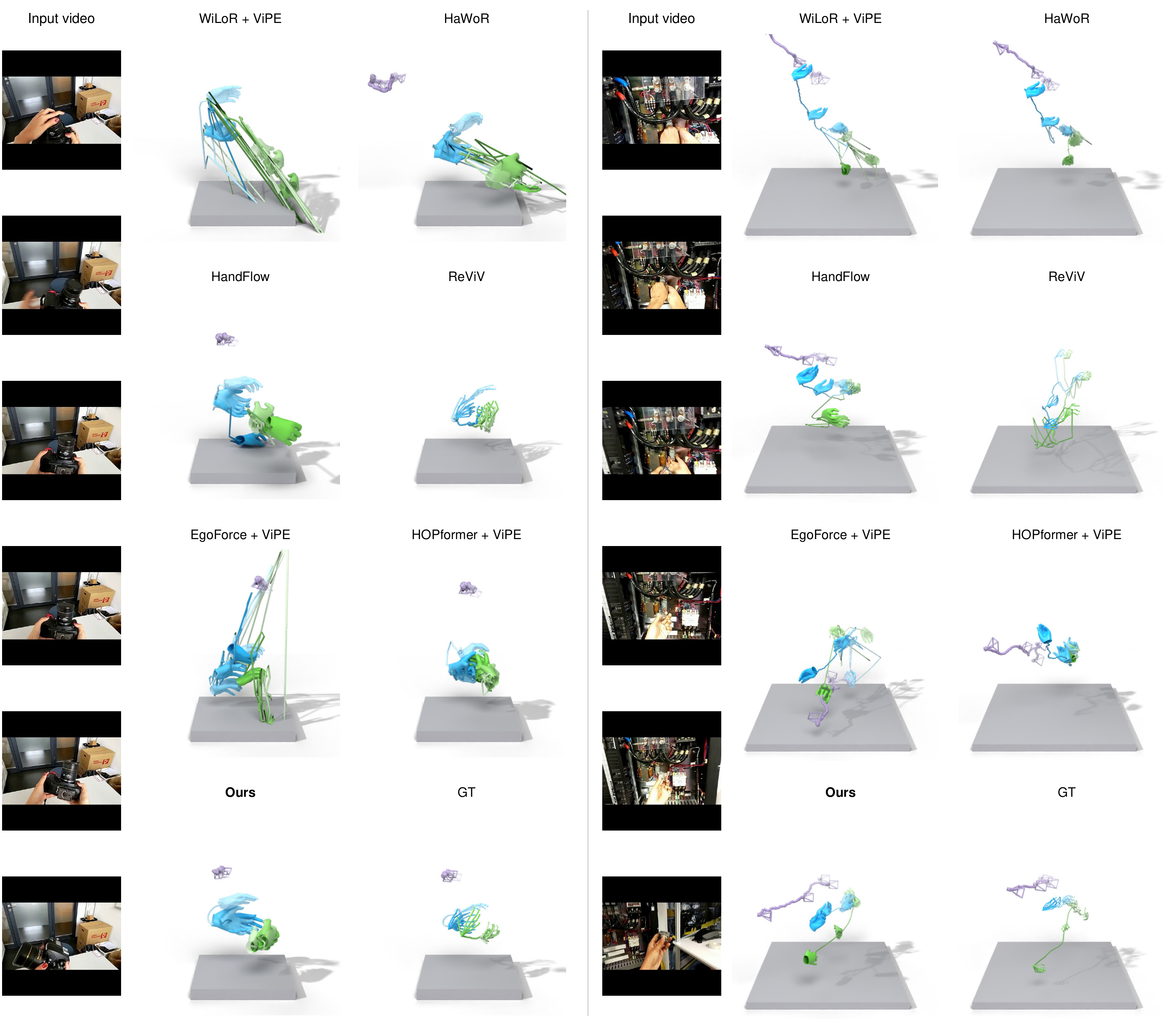}
\vspace{-5pt}
\caption{\textbf{Qualitative comparison on HoloAssist.} Four sequences are shown, two per row.}
\label{fig:qual_holoassist}
\end{figure}

\begin{figure}[p]
\centering
\includegraphics[width=\linewidth,height=.43\textheight,keepaspectratio]{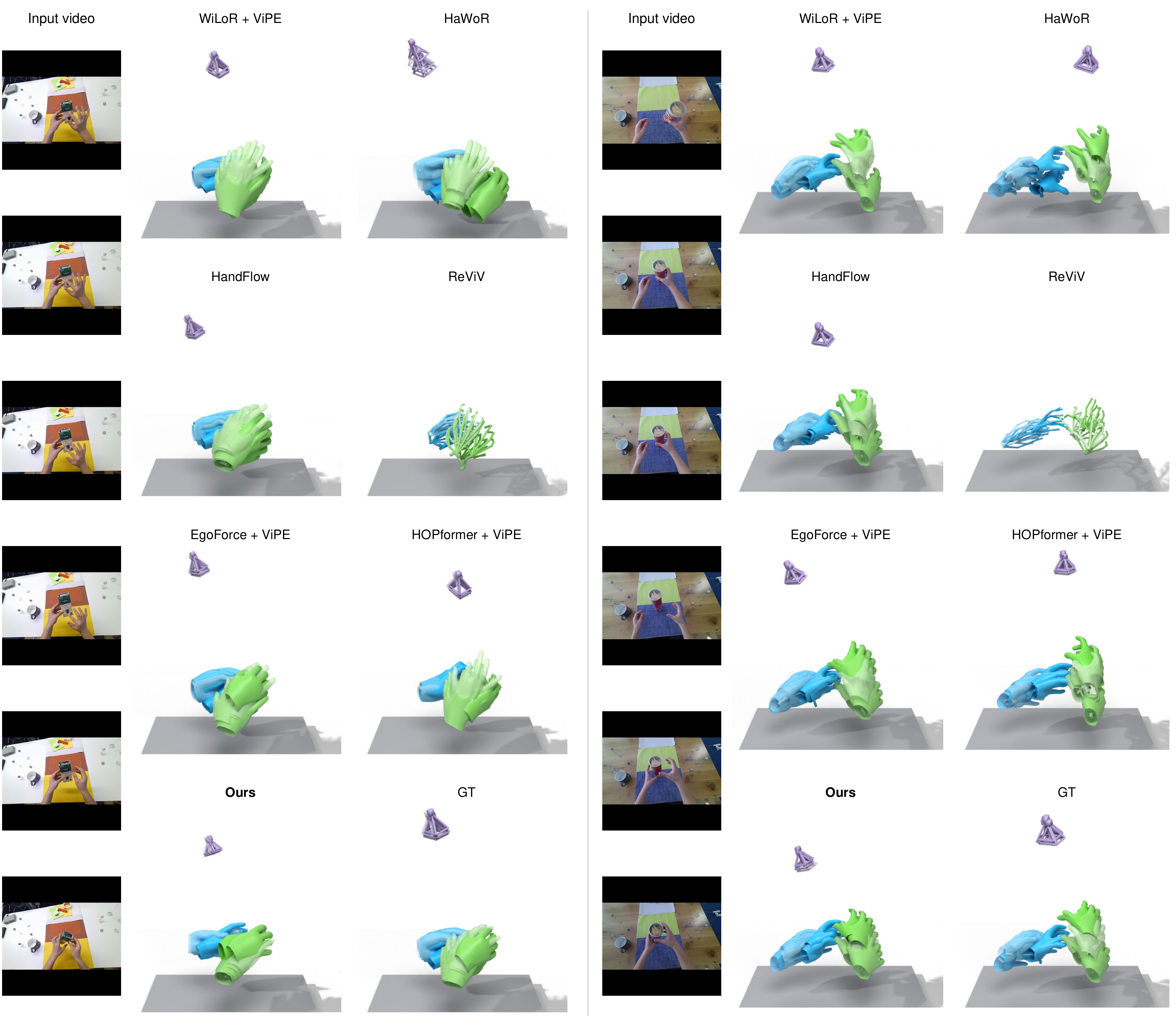}\par
\vspace{10pt}
\includegraphics[width=\linewidth,height=.43\textheight,keepaspectratio]{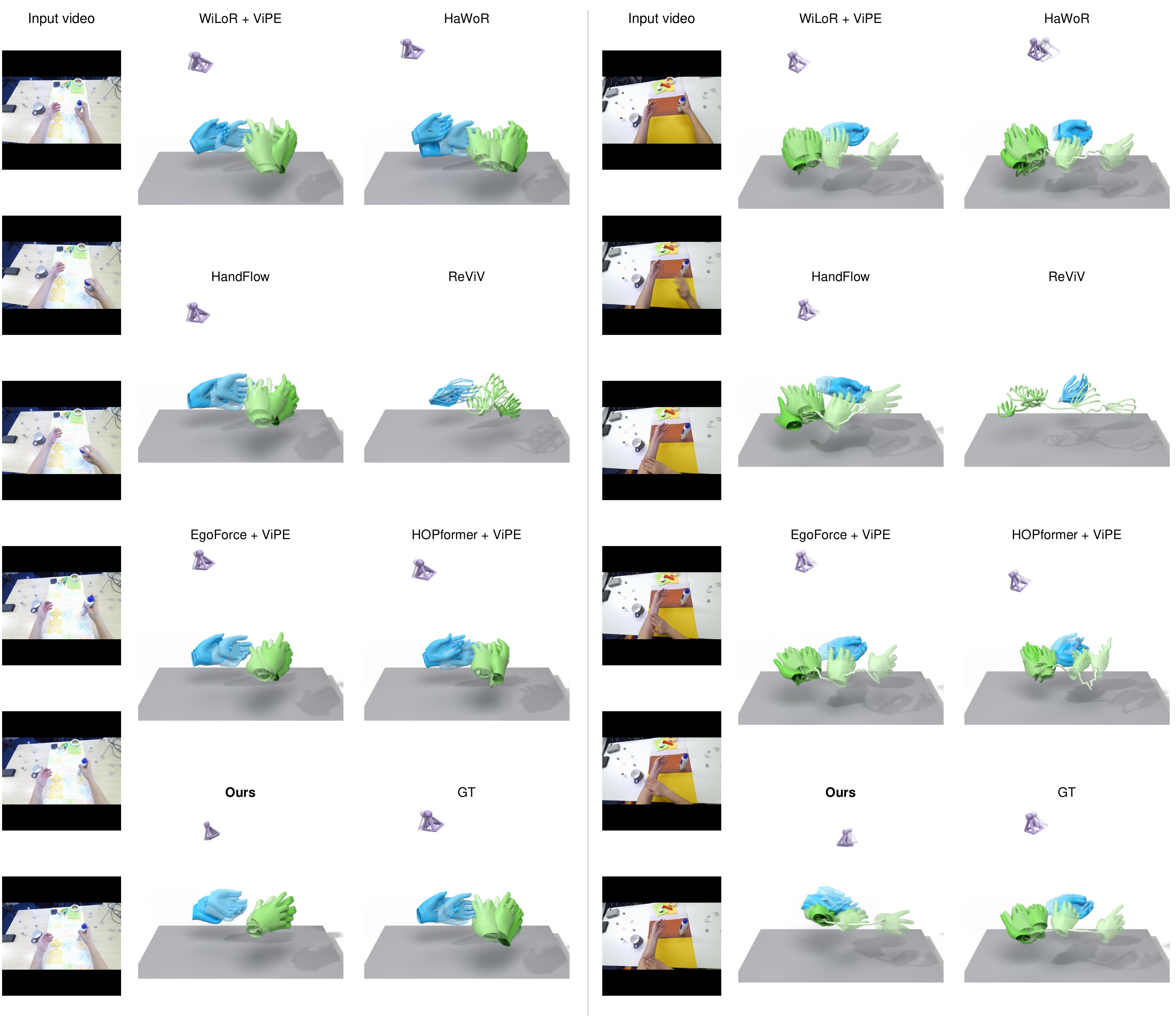}
\vspace{-5pt}
\caption{\textbf{Qualitative comparison on H2O.} Four sequences are shown, two per row.}
\label{fig:qual_h2o}
\end{figure}

\begin{figure}[p]
\centering
\includegraphics[width=\linewidth,height=.43\textheight,keepaspectratio]{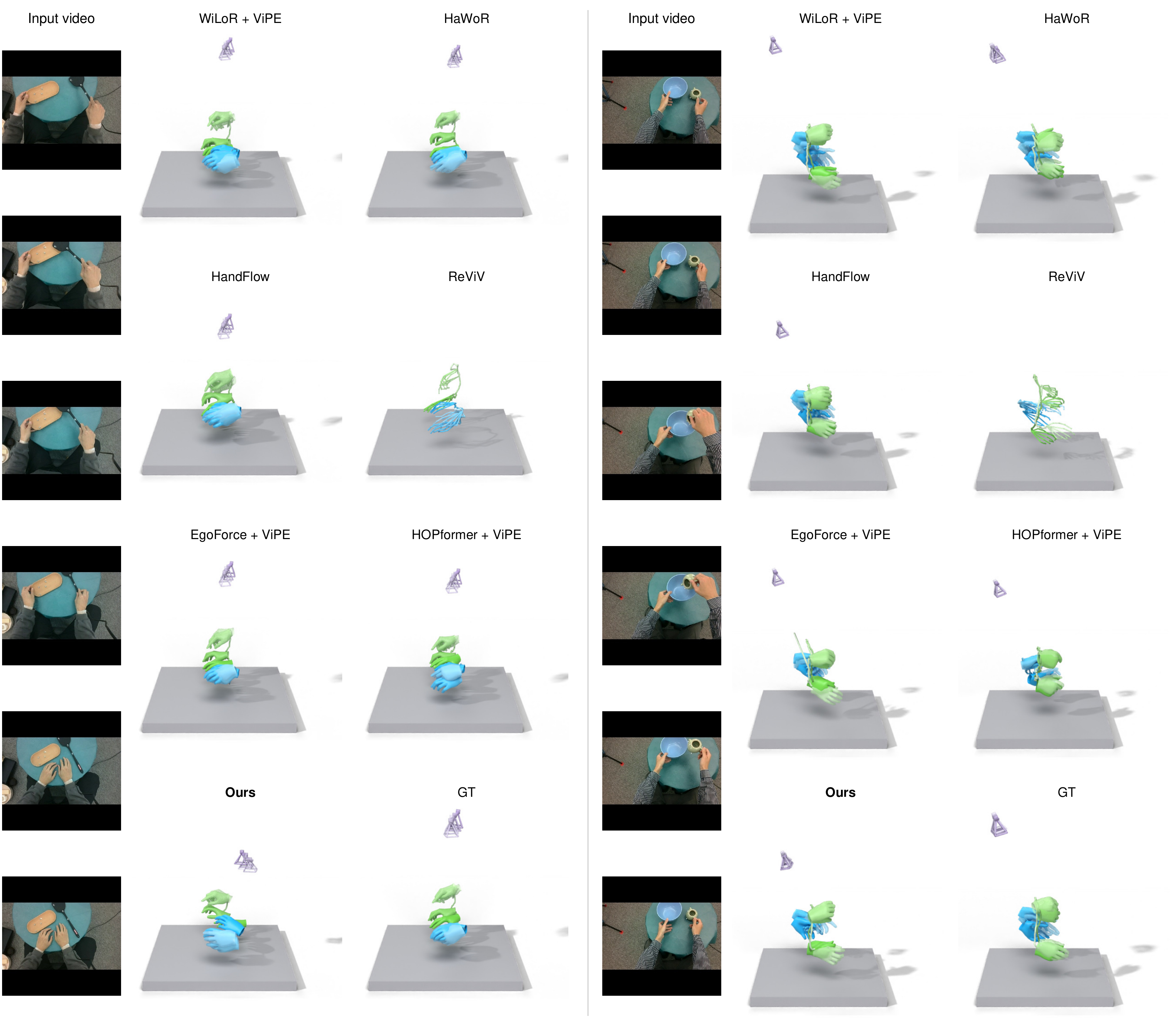}\par
\vspace{10pt}
\includegraphics[width=\linewidth,height=.43\textheight,keepaspectratio]{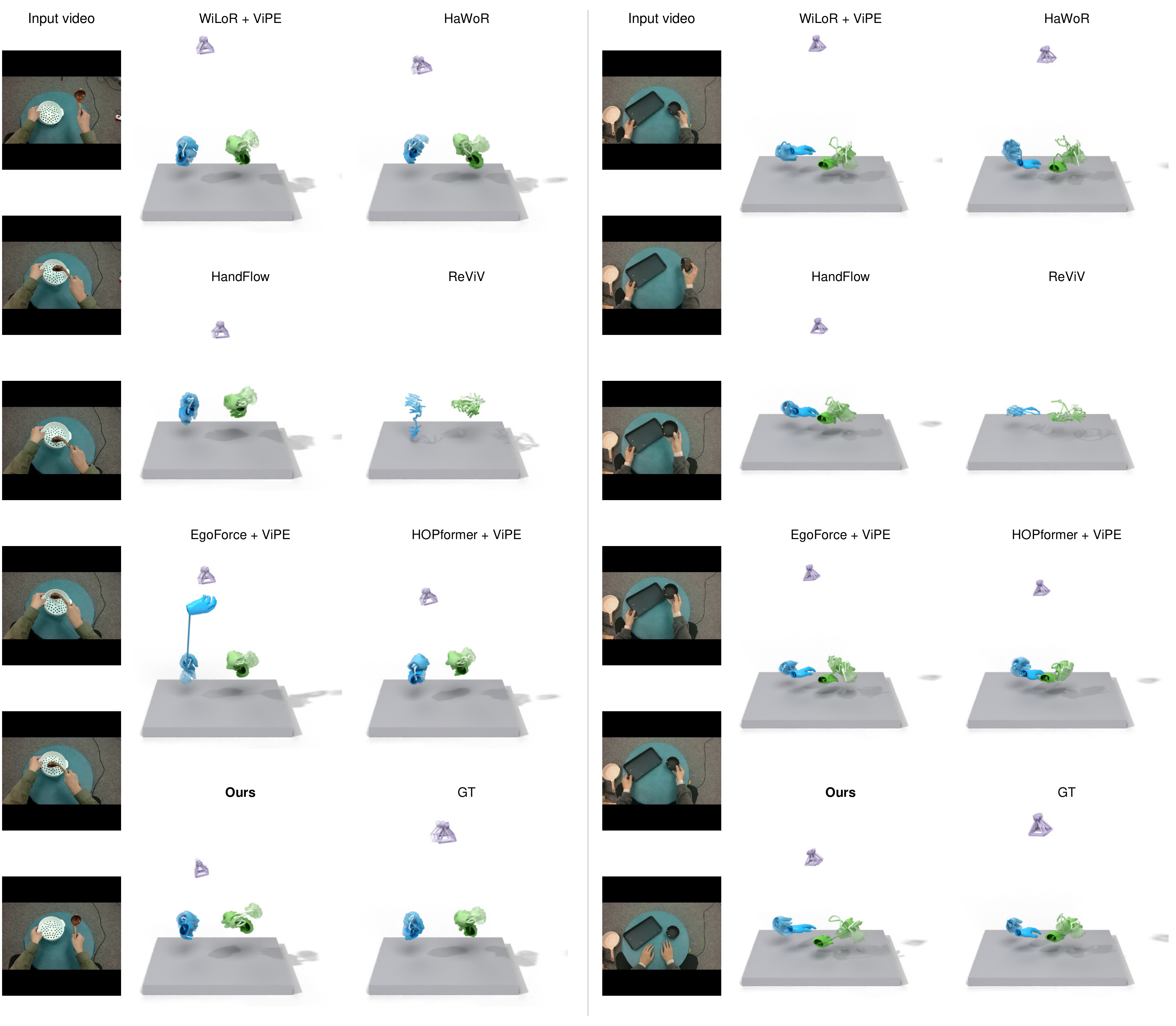}
\vspace{-5pt}
\caption{\textbf{Qualitative comparison on TACO.} Four sequences are shown, two per row.}
\label{fig:qual_taco}
\end{figure}

Figs.~\ref{fig:qual_hot3d_full} through~\ref{fig:qual_taco} show additional examples on HOT3D, ARCTIC,
HoloAssist, H2O, and TACO. All methods in each sequence share one viewpoint and display scale.
For each method, one $\mathrm{SE}(3)$ transform fitted to both hands is also applied
to the camera, without scale correction. Green and blue curves trace the right and left wrists,
and purple frustums indicate the camera. HoloAssist GT and ReViV use 21-joint skeletons; gray
platforms are rendering aids.

\end{document}